\RequirePackage{fix-cm}
\documentclass[twocolumn]{svjour3}          
\smartqed  

\usepackage{graphicx}
\usepackage{amsmath}
\usepackage{amssymb}
\usepackage{booktabs}
\usepackage[normalem]{ulem}
\usepackage{tabularx}
\usepackage{tikz}
\usepackage{makecell}
\usepackage{caption}
\usepackage{subcaption}
\usepackage{multirow}
\usepackage{ulem}
\usepackage[misc,geometry]{ifsym}
\def\etal{\textit{et al}.}
\def\ie{\textit{i.e.}}
\def\eg{\textit{e.g.}}

\newcommand{\zy}[1]{\textcolor{black}{#1}}
\usepackage[breaklinks=true,letterpaper=true,colorlinks,bookmarks=false]{hyperref}

\journalname{International Journal of Computer Vision}
\begin{document}\sloppy

\title{Single-View View Synthesis with Self-Rectified Pseudo-Stereo
}



\author{Yang Zhou \and Hanjie Wu \and Wenxi Liu \and Zheng Xiong \and Jing Qin \and Shengfeng He \Letter}


\institute{Yang Zhou, Hanjie Wu, Zheng Xiong \at
           School of Computer Science and Engineering, South China University of Technology \\
              \email{matrixGle19@gmail.com, cshanjiewu@gmail.com, pandadreamer21@gmail.com}           
           \and
           Wenxi Liu \at
           College of Mathematics and Computer Science, Fuzhou University \\
              \email{wenxi.liu@hotmail.com}
           \and
           Jing Qin \at
           Department of Nursing, Hong Kong Polytechnic University \\
              \email{harry.qin@polyu.edu.hk}
           \and
           Shengfeng He (\Letter) \at
           School of Computing and Information Systems, Singapore Management University\\
              \email{shengfenghe@smu.edu.sg}
}

\date{Received: date / Accepted: date}

\maketitle

\begin{abstract}
  Synthesizing novel views from a single view image is a highly ill-posed problem. We discover an effective solution to reduce the learning ambiguity by expanding the single-view view synthesis problem to a multi-view setting. Specifically, we leverage the reliable and explicit stereo prior to generate a pseudo-stereo viewpoint, which serves as an auxiliary input to construct the 3D space. In this way, the challenging novel view synthesis process is decoupled into two simpler problems of stereo synthesis and 3D reconstruction. In order to synthesize a structurally correct and detail-preserved stereo image, we propose a \zy{self-rectified stereo synthesis} to amend erroneous regions in an identify-rectify manner. Hard-to-train and incorrect warping samples are first discovered by two strategies, 1) pruning the network to reveal low-confident predictions; and 2) bidirectionally matching between stereo images to allow the discovery of improper mapping. These regions are then inpainted to form the final pseudo-stereo. With the aid of this extra input, a preferable 3D reconstruction can be easily obtained, and our method can work with arbitrary 3D representations. Extensive experiments show that our method outperforms state-of-the-art single-view view synthesis methods and stereo synthesis methods.

\end{abstract}

{\section{Introduction}\label{sec:introduction}}

\begin{figure*}[t]
    \centering
    \begin{subfigure}{.194\linewidth}
        \centering
        \includegraphics[width=.99\linewidth]{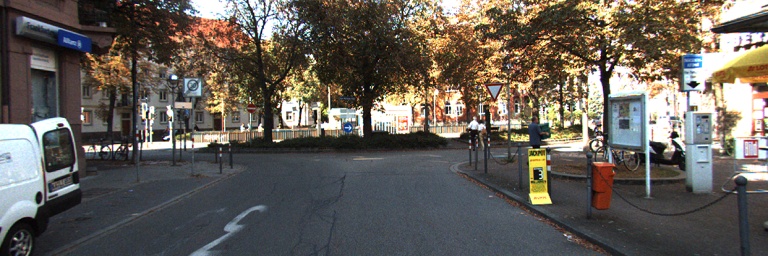}\vspace{1mm}
        \includegraphics[width=.99\linewidth]{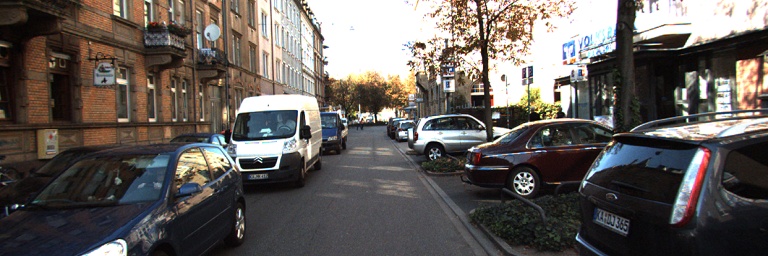}\vspace{1mm}
        \includegraphics[width=.99\linewidth]{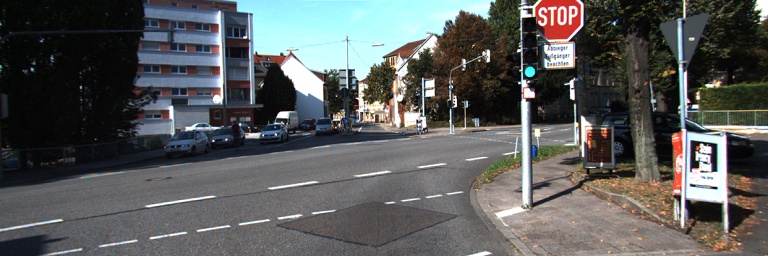}\vspace{1mm}
        \includegraphics[width=.99\linewidth]{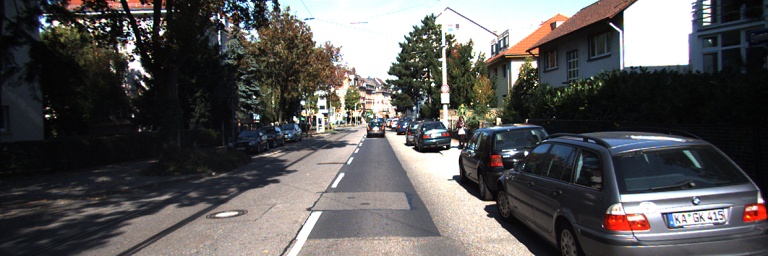}\vspace{1mm}
        \caption{Input}
    \end{subfigure}
    \begin{subfigure}{.194\linewidth}
        \centering
        \includegraphics[width=.99\linewidth]{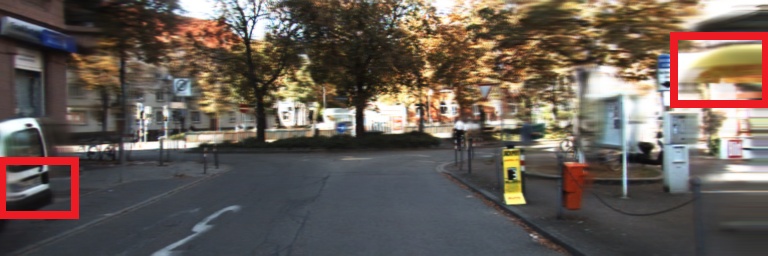}\vspace{1mm}
        \includegraphics[width=.99\linewidth]{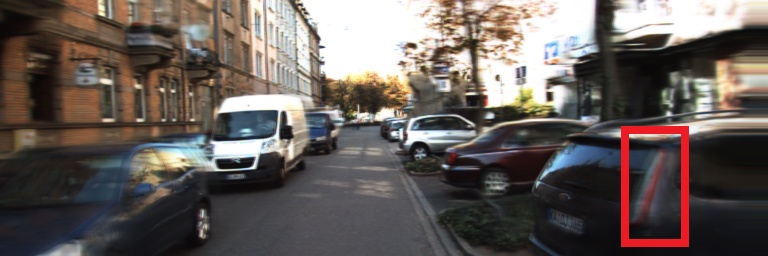}\vspace{1mm}
        \includegraphics[width=.99\linewidth]{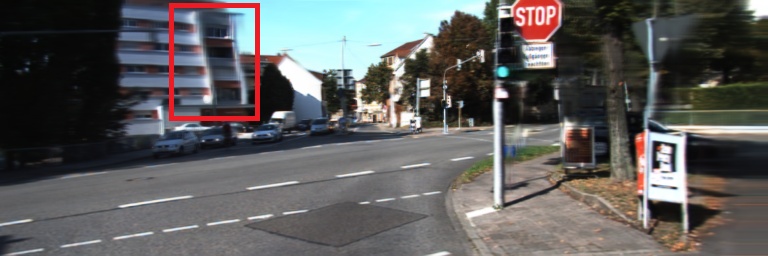}\vspace{1mm}
        \includegraphics[width=.99\linewidth]{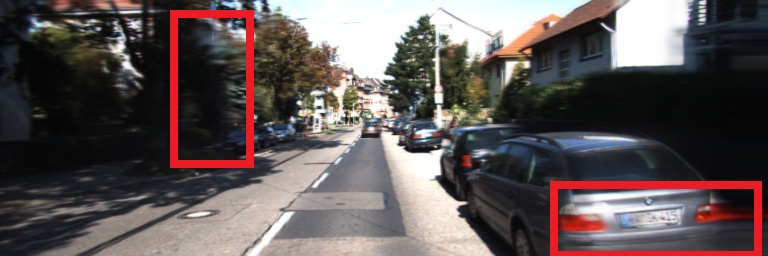}\vspace{1mm}
        \caption{MINE~\cite{li2021mine}}
        \label{teaser:sota}
    \end{subfigure}
    \begin{subfigure}{.194\linewidth}
        \centering
        \includegraphics[width=.99\linewidth]{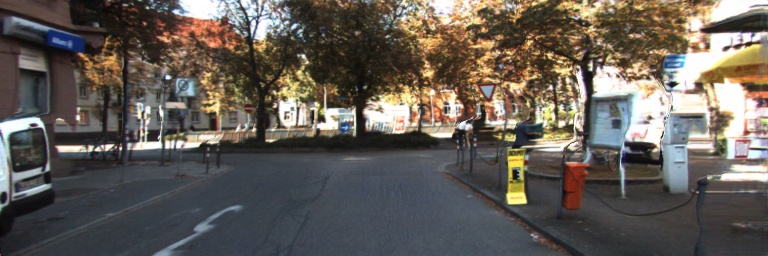}\vspace{1mm}
        \includegraphics[width=.99\linewidth]{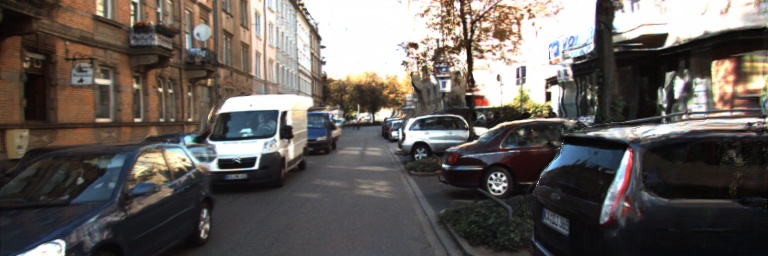}\vspace{1mm}
        \includegraphics[width=.99\linewidth]{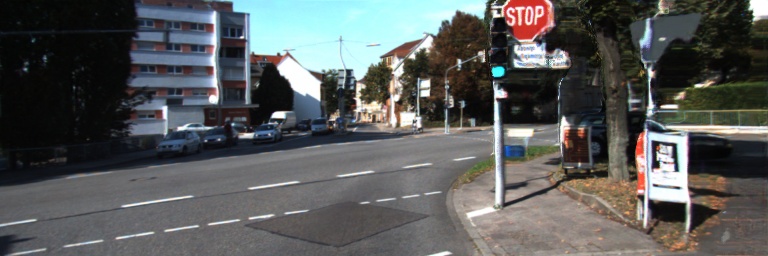}\vspace{1mm}
        \includegraphics[width=.99\linewidth]{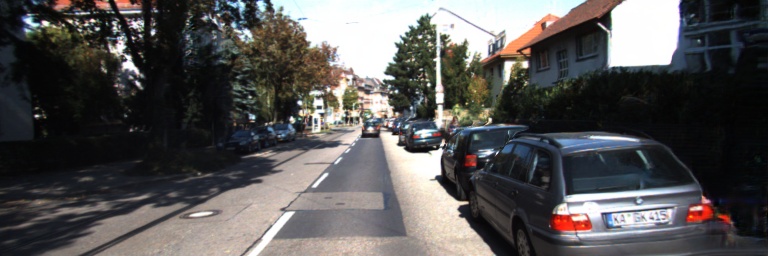}\vspace{1mm}
        \caption{Pseudo Stereo}
        \label{teaser:stereo}
    \end{subfigure}
    \begin{subfigure}{.194\linewidth}
	\centering
        \includegraphics[width=.99\linewidth]{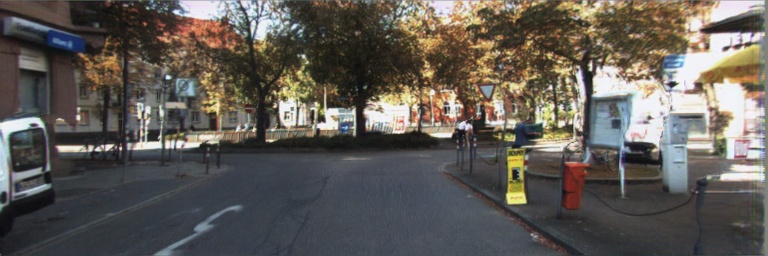}\vspace{1mm}
        \includegraphics[width=.99\linewidth]{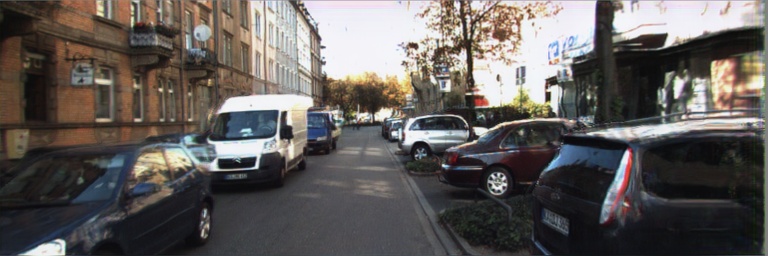}\vspace{1mm}
        \includegraphics[width=.99\linewidth]{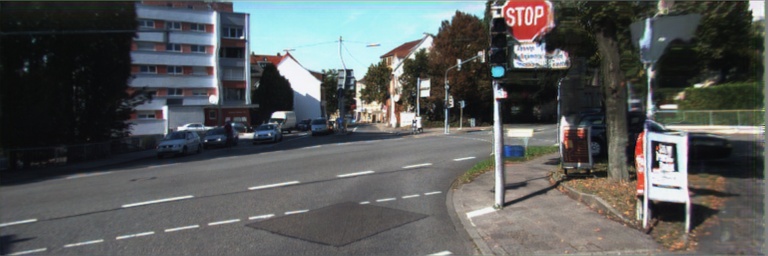}\vspace{1mm}
        \includegraphics[width=.99\linewidth]{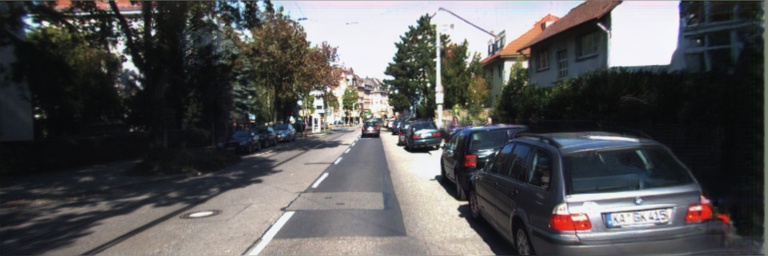}\vspace{1mm}
	\caption{Ours}
	\label{teaser:ours}
	\end{subfigure}
    \begin{subfigure}{.194\linewidth}
	\centering
        \includegraphics[width=.99\linewidth]{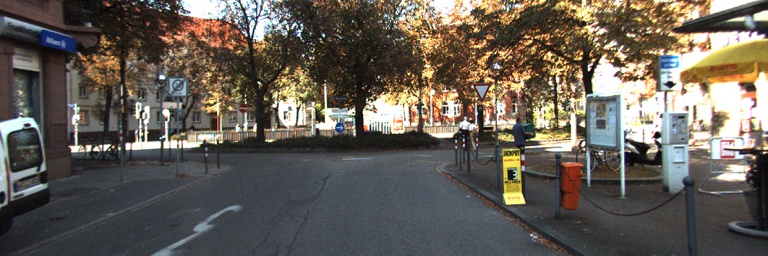}\vspace{1mm}
        \includegraphics[width=.99\linewidth]{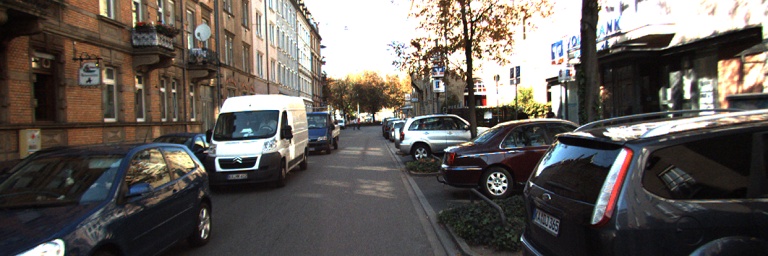}\vspace{1mm}
        \includegraphics[width=.99\linewidth]{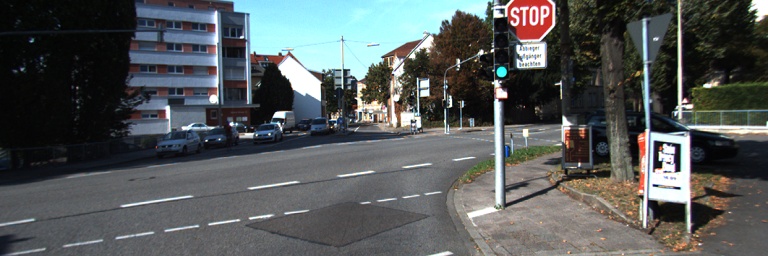}\vspace{1mm}
        \includegraphics[width=.99\linewidth]{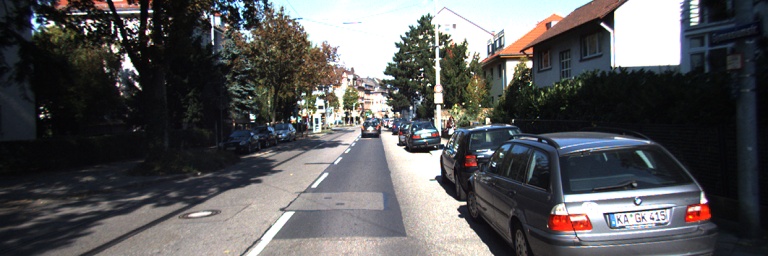}\vspace{1mm}
	\caption{GT}
  \end{subfigure}
  \caption{Existing methods like MINE~\cite{li2021mine} suffer from severely ambiguous 3D reconstruction from a single view, and thus fail to produce visual plausible synthesis results. \zy{For instance, the shape of building is distorted, and the license plate blurry.} In this work, we relieve the ambiguity by decoupling this problem into two subproblems, pseudo stereo synthesis and 3D reconstruction. Reliable novel view synthesis can be easily obtained with the aid of our authentic rectified pseudo-stereo.}
  \label{fig:teaser}
\end{figure*}

3D photography enables the display of a scene from arbitrary viewpoints, bringing immersive experiences like virtual or augmented reality. To create a 3D world from limited input images, the problem of 3D photography shifts to the synthesis of novel views by reasoning and understanding the occlusion and 3D structure of the scene.

Due to the environmental constraints, users may take various types of input photos which can lead to substantially different reconstruction algorithms. It is natural that the more images are captured from the same scene, the more reliable reconstruction can be obtained. As a consequence, multi-view inputs, ranging from stereo pair~\cite{zhou2018stereo} to twelve images from a camera array~\cite{Flynn2019DeepViewVS}, are required to construct the 3D representations, like multiplane image (MPI)~\cite{Szeliski2004StereoMW}. Despite their preferable reconstruction, the strict input requirement prevents non-professional users from sharing their real-life environments in the form of virtual reality.

Recent advances show that 3D photography can be achieved using a single-view image. According to the 3D representations they used, their methods are tailored to remedy the problems in that representation. For instance, layer depth image (LDI) based methods~\cite{Shih20203DPU,kopf2020one} aim to inpaint occlusions between layers, while MPI-based methods~\cite{Tucker2020SingleViewVS,li2021mine} focus on continuing the discrete 3D space. Regardless of the 3D representations, we argue that constructing a 3D space solely from a single-view image is ill-posed as the direct mapping between two ends is highly disproportionate (see Figure~\ref{teaser:sota}). As a consequence, extra prior knowledge is much needed to alleviate this ambiguity.

In this paper, instead of designing a preferable 3D presentation, we turn to a novel perspective of augmenting the input space before reconstructing the scene. Comparing to generating arbitrary viewpoints, we observe that images captured by the same stereo camera have a fixed baseline, which erases most of uncertain factors during synthesis. We, therefore, expand the single-view view synthesis problem to a pseudo multi-view regime by virtue of explicit stereo prior, which significantly eases the difficulty of 3D reconstruction.

In particular, we concentrate on how to generate a faithful right view given the left one. To this end, we perform stereo synthesis in two steps, stereo warping and \zy{synthesis rectification}. The former step obtains a preliminary result by learning a stereo flow that warps the left view to the right one. Constructing the right view with pixel-level warping has a clear advantage of detail preservation, while its downside is the distorted structure, especially for thin objects. Thus, the latter step identifies and rectifies the inconsistencies in the warping result. Detecting inconsistencies is non-trivial and we have two strategies. First, we discover hard-to-train samples in the stereo flow network that produce low-confident predictions using network pruning. The pruned network infers differently on inconsistent regions while remaining the same on the others. Second, we introduce a bidirectional matching model to identify erroneous regions of the image, as a good pixel-pair should be consistently matched in either ways. Those found regions are then inpainted with authentic image content, producing a high-quality stereoscopic image. In the end, with the aid of our stereo prior, the reconstructed 3D space is robust and accurate. More importantly, the proposed method is representation-independent and can work with arbitrary 3D representations. Extensive experiments demonstrate the superior performance over state-of-the-art single-view view synthesis as well as stereo synthesis methods.

Our contributions can be summarized as follows:
\begin{itemize}
\item We discover that the single-view view synthesis problem can be effectively decomposed into two subproblems with extra pseudo prior knowledge. This idea provides new insights in advancing the performance in this field.
\item We tailor two self-rectification methods for stereo synthesis, which integrates the advantages of warping and inpainting models, resulting in structurally correct and detail-preserved stereo images.
\item We set a new state-of-the-art in the areas of single-view view synthesis as well as stereo synthesis.
\end{itemize}

\section{Related Work}\label{sec:related}
\textbf{Multi-view view synthesis.} Traditional methods are proposed based on image rendering techniques \cite{Hedman2018BF,Penner2017Soft3R,Sinha2012ImagebasedRF} and multi-view geometry principles \cite{Fitzgibbon2005ImageBasedRU,Kopf2013ImagebasedRI,Karras2019ASG,Seitz2006ACA,Debevec1998EfficientVI,debevec1996modeling}. These methods model the scene and reason about novel views from multiple images along with depth images as inputs \cite{Chaurasia2013DepthSA,Penner2017Soft3R}. On the other hand, some deep learning-based methods \cite{Novotn2019PerspectiveNetAS,Meshry2019NeuralRI,MartinBrualla2018LookinGoodEP,Hedman2018BF,Choi2019ExtremeVS,Aliev2020NeuralPG} {utilize the powerful generation capability of deep neural networks with depth maps as auxiliary inputs} to produce view synthesis results, without explicitly modeling the 3D scene structure.
Recent methods \cite{Jantet2009IncrementalLDIFM,Flynn2019DeepViewVS,Srinivasan2019PushingTB,zhou2018stereo,Mildenhall2020NeRFRS,Park2020DeformableNR} follow the pipeline of first modeling the scene and then rendering the novel-view image according to the estimated 3D representation.
Based on this pipeline, different types of 3D representations can be estimated including layered depth image (LDI) \cite{Jantet2009IncrementalLDIFM}, multi-plane images (MPI)~\cite{Flynn2019DeepViewVS,Srinivasan2019PushingTB,zhou2018stereo}, and neural radiance fields \cite{Mildenhall2020NeRFRS,Park2020DeformableNR}. Although multi-image based methods can preferably synthesize novel views, it is not flexible enough to promote to non-professional users.

\textbf{Single-view view synthesis.} Single-image based view synthesis method is often formulated as an image-to-image translation task~\cite{Zhou2016ViewSB,Tatarchenko2016Multiview3M,Sun2018MultiviewTN,Park2017TransformationGroundedIG,Kulkarni2015DeepCI,Chen2016InfoGANIR}. There are also a large body of works that first attempt to predict 3D representations and then synthesize the novel-view image accordingly. For example, \cite{Tulsiani2018Layerstructured3S,Shih20203DPU} apply LDI while~\cite{Tucker2020SingleViewVS,li2021mine} employ MPI as the 3D representation. Recent works \cite{Tucker2020SingleViewVS,li2021mine} construct the MPI representation from a single input image first, and then produce a novel-view image by rendering techniques such as homograph warping and integration of the planes.
\cite{jampani2021slide} proposes a soft layering strategy with depth-aware inpainting to preserve more details in novel views.
However, the information that a single image can provide is limited, since the associations of objects may not be observed from other views. In order to alleviate the ambiguity caused by constructing a 3D scene representation from a single image, we use the explicit stereo prior to generate a pseudo-stereo viewpoint for better generating a 3D representation like Figure ~\ref{teaser:ours}.

\begin{figure*}[t]
	\begin{center}
		\includegraphics[width=\linewidth]{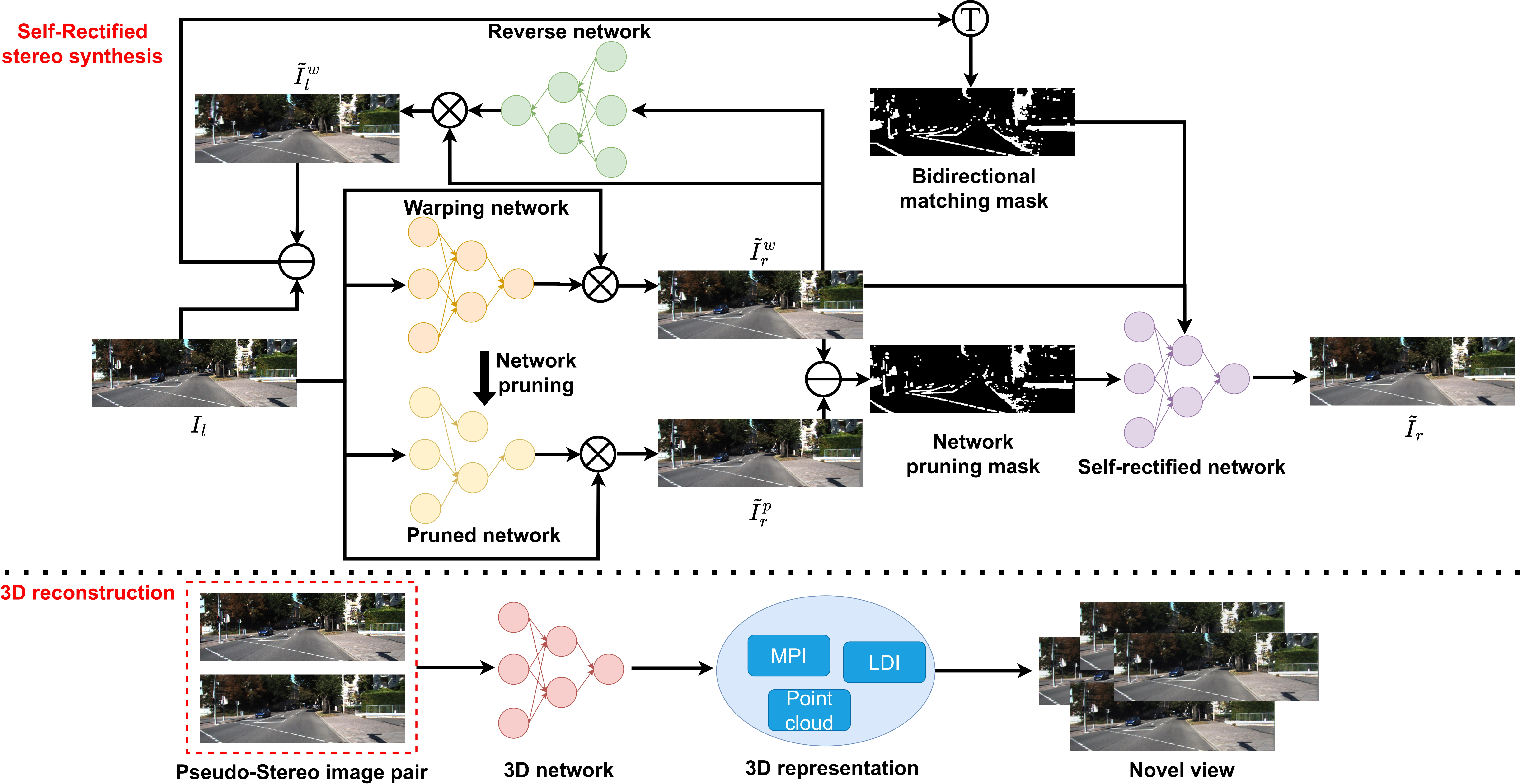}
	\end{center}
    \caption{Illustration of the pipeline of our method. It first learns the stereo flow to produce a warped right view by a stereo warping network in a self-supervising manner.
    Then, the preliminary result is fed to the reverse network and pruned network for locating the structurally incorrect regions.
	Note that the regions found from reverse network will be mapped into the target view based on the track-back function.
    These regions will be filled with realistic content by the \zy{synthesis rectification} network to obtain the pseudo right-view image, and thus pair with the input left-view image as stereo pair.
    With such a pseudo-stereo pair, we can reconstruct 3D representation of the scene by 3D reconstruction module. Since our method is representation-independent, 3D representation can be arbitrary, \eg, MPI, LDI, and so forth. On this basis of reconstructed 3D information, we can synthesize novel-view images.}
	\label{fig:pipeline}
\end{figure*}

\textbf{Stereo synthesis.} 3D stereo images can be displayed on VR headsets and glasses, and stereo synthesis can be considered as a special case of novel view synthesis. \cite{Szeliski2004StereoMW,zhou2018stereo} perform stereo magnification that synthesizes novel stereo views with stereo images as input using the MPI representation.
Xie \etal \cite{Xie2016Deep3DFA} investigate the problem of generating the corresponding right view image from the input left view image, but it fails to model the object occlusion relationships. Both \cite{Godard2017UnsupervisedMD} and \cite{luo2018single} aim to accomplish the depth estimation task, given a single image as input. They first generate a stereo pair from the input, and then perform depth estimation based on the stereo pair. \cite{Cun2019DepthAssistedFR,watson2020learning} take a reverse pipeline to first estimate disparity from a single image then synthesize stereo pairs. These methods demonstrate that stereo prior knowledge can provide more effective information for downstream tasks. However, these methods treat all matching samples equally, and fails to remedy the influences brought by bad sample pairs. We explicitly discover and recover those bad samples, producing preferable stereo synthesis results.

\textbf{Network pruning.} Network pruning is one of network compression approaches to remove the redundant branches while sacrificing a small amount of accuracy \cite{Liu2017LearningEC,Li2017PruningFF,Frankle2019TheLT}. Specifically, Hooker \etal \cite{hooker2019compressed} find out that pruning a classification network has an uneven impact on samples from different categories. The most affected samples belong to the long-tailed, hard-to-learn categories in the data distribution. Inspired by the findings, \cite{Wang2021TroubleshootingBI} proposes a set called self-competitor to improve the robustness of the blind image quality assessment model. \cite{Jiang2021SelfDamagingCL} presents a dynamic self-competitor model by pruning the full model and comparing its original complete features with the features of the pruned model, which can increase the weights of difficult samples in contrast loss, and thus achieve rebalancing of different sample losses. Inspired by the unique and effective property of network pruning, we utilize it to discover unreliable warping results.

\section{Overview}

{In this work, our goal is to synthesize a novel-view image from a single image.
Instead of learning a preferable 3D presentation of the target scene, we propose to expand the single-view view synthesis problem to a pseudo multi-view regime, \ie, generating a faithful right view given the input single image as the left view, by virtue of the explicit stereo prior, so as to obtain a robust and accurate 3D reconstruction to work with arbitrary 3D representations.}

{In particular, we perform stereo synthesis in two steps, stereo warping and \zy{synthesis rectification}. The former step obtains a preliminary result by learning a stereo flow that warps the left view to the right one. The latter step identifies and rectifies the inconsistencies in the warping result. As follows, we will elaborate the details of stereo synthesis and 3D reconstruction. The pipeline of our method is depicted in Figure ~\ref{fig:pipeline}.}

\section{\zy{Self-Rectified Stereo Synthesis}}\label{sec:method_stereo}
In this section, we describe our self-rectified stereo image synthesis approach.
It takes an single left-view image ${I_l}$ as input, to generate a right-view image $~\tilde{I}_r$ that forms a stereo image pair with ${I_l}$, \ie,:
\begin{equation}
	\tilde{I}_r = f_S(I_l),
\end{equation}
where $f_S(\cdot)$ denotes a mapping function that maps the left-view to the right-view (\ie, self-rectified stereo synthesis network in our case).
Specifically, we perform stereo synthesis in two steps: self-supervised stereo warping and \zy{synthesis rectification}, \ie,:
\begin{equation}
    f_S(\cdot) \equiv f_R(f_W(\cdot)),
\end{equation}
where $f_R(\cdot)$ and $f_W(\cdot)$ refer to the \zy{synthesis rectification} and stereo warping module, respectively.

\subsection{Self-supervised Stereo Warping}
Since most of the scene in $I_l$ appears in $\tilde{I}_r$, it is intuitive to learn a stereo flow for warping $I_l$ to achieve the preliminary result of right-view image $\tilde{I}_r^w$, as below.
\begin{equation}
	\tilde{I}_r^w = f_W(I_l).
\end{equation}
Instead of generating the right-view image directly, stereo warping tends to produce better fine details.
Nevertheless, the ground-truth flow is unknown, so we propose a self-supervised stereo warping model,
$f_W(\cdot)$, to learn a stereo flow in order to strengthen the warped image.

Empirically, the stereoscopic images that are captured by the same stereo cameras have a fixed baseline. This fact can be used to erase other uncertainties, so we allow the stereo flow to be horizontal displacement only. {Regarding the network structure, we utilize an encoder-decoder structure as the backbone of stereo warping network $f_W(\cdot)$ for feature extraction
and the flow prediction module following \cite{ilg2017flownet}.}

\begin{figure}[t]
	\centering
	\setlength{\fboxsep}{0.01mm}
	\begin{subfigure}[t]{7.5cm}
		\fbox{\includegraphics[width=0.99\linewidth]{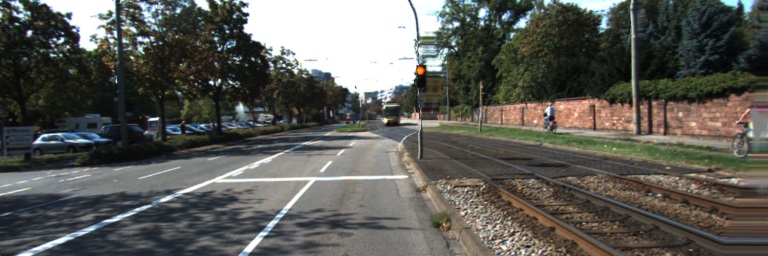}}
		\caption{Warped image}
	\end{subfigure}
	\begin{subfigure}[t]{7.5cm}
		\fbox{\includegraphics[width=0.99\linewidth]{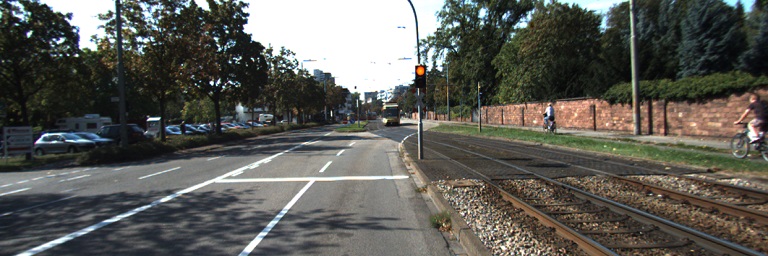}}
		\caption{GT}
	\end{subfigure}

	\begin{subfigure}[t]{7.5cm}
		\fbox{\includegraphics[width=0.99\linewidth]{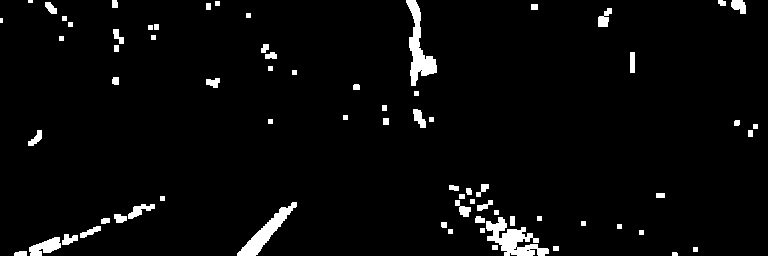}}
		\caption{Pruning mask}
	\end{subfigure}
	\begin{subfigure}[t]{7.5cm}
		\fbox{\includegraphics[width=0.99\linewidth]{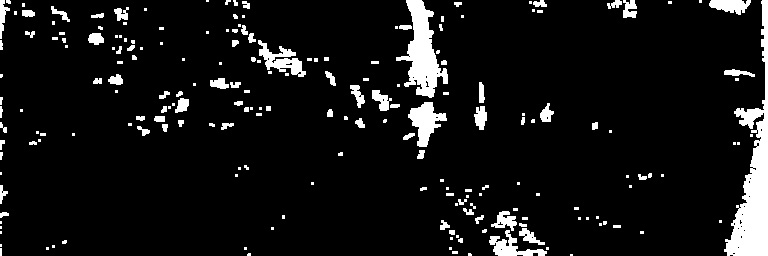}}
		\caption{Bidirectional mask}
	\end{subfigure}
	\caption{Inconsistent region identified by our \zy{synthesis rectification}.}
	\label{fig:mask}
\end{figure}

Although constructing the right view with pixel-level warping has obvious advantage for preserving details from $I_l$ to $\tilde{I}_r$, its downside is the distorted structure (especially for thin objects), and possible missing content in the right view.
These issues may cause erogenous regions in the warped image.
{As illustrated in the first row of Figure \ref{fig:mask}, distortions appear around the thin objects, like the signal lamppost. And the region at right is blurry, which suffers a lot from the missing content.}
As it is non-trivial to identify these inconsistent regions, we propose two \zy{synthesis rectification} strategies, \ie, pruning-based and bidirectional match rectification. We will elaborate them in the following.

\subsection{Pruning-based Rectification}
Our first \zy{synthesis rectification} strategy is pruning-based rectification.
As demonstrated in \cite{hooker2019compressed}, network pruning has a non-uniform impact on the input samples. Concretely, for hard-to-train samples, referred as dubbed pruning identified exemplars (PIEs) in \cite{hooker2019compressed}, pruned network will produce low confident predictions, comparing to simple samples. This is because branches producing those predictions are typically pruned to balance the performance and model size.
It inspires our pruning-based rectification module that is enabling to detect the erroneous regions of warped images, \ie, PIE regions, before rectifying them.
By pruning the trained network, it will lead to non-uniform influence on all pixels. The hard-to-warp regions are impacted more while the others are less changed.

To accomplish this, we perform magnitude-based pruning for our stereo warping network $f_W(\cdot)$. In particular, we adopt a simple pruning strategy that removes the smallest-magnitude weights of the network under a percentage $p$.
We then feed $I_l$ into the resulting pruned networks $f_W^p(\cdot)$ to obtain the ``pruned'' right image $\tilde{I}_r^p$.
By comparing the outcomes of the original network and pruned network, $\tilde{I}_r^w$ and $\tilde{I}_r^p$, we can localize the PIEs regions via their discrepancy. For those image regions with large discrepancy, they are low-confident or uncertain regions for the stereo warping network $f_W(\cdot)$ to handle.
{As shown in the third row of Figure  \ref{fig:mask}, pruning-based rectification module is adept at finding the intractable regions of warping,
particularly the thin objects like the curb and signal lamppost.}
With inconsistent regions being found, we inpaint them with faithful image contents.
Hence, this process is formulated as follows:
\begin{equation}
	\tilde{I}_r^p = f_W^p(I_l),
\end{equation}
\begin{equation}
	\vspace{-50mm}
	\delta^p = \Delta(\tilde{I}_r^p, \tilde{I}_r^w),
\end{equation}
where $\Delta(\cdot, \cdot)$ refers to the function for computing pixelwise difference, {and $\delta^p$ represents the difference of $\tilde{I}_r^p$ and $\tilde{I}_r^w$, \ie, confidence map of $I_r^w$. }
Specifically, $\Delta(\cdot, \cdot)$ first calculate the difference and then carry out normalization on the difference, that is, $\delta^p \in [0, 1]$.

\subsection{Bidirectional Matching Rectification}

Our second \zy{synthesis rectification} strategy is bidirectional matching rectification.
Its motivation lies in the fact that good pixel-pair should be consistently matched in either ways, while erroneous pixels in warped images can hardly be warped back to the original image.

\zy{Insted of the typical left-right consistency check carried out in the disparity space, we perform matching check in pixel space in favor of avoiding the smoothness characteristics of disparity, which leads to more accurate results.}
In specific, we present another network $f_B(\cdot)$ to learn the backward stereo flow that warps $\tilde{I}_r^w$ back to the left view, i.e., $\tilde{I}_l^w$.
For those inconsistent pixels, it results in erroneous regions of $\tilde{I}_l^w$.
By means of bidirectional matching, the large discrepancy between $\tilde{I}_l^w$ and $I_l$ implies the irreversible pixels,
and we can trace them back to the view of $\tilde{I}_r^w$.
{The last row in Figure  \ref{fig:mask} demonstrates that besides the distorted structures, bidirectional matching rectification is able to identifies those missing regions of right view accurately as well.
This brings benefits to improve the effect of inpainting method.}
Similar to pruning-based rectification module, we have the following process:
\begin{equation}
	\tilde{I}_l^w = f_B(\tilde{I}_r^w),
\end{equation}
\begin{equation}
	\delta^b = T(\Delta(\tilde{I}_l^w, I_l)),
\end{equation}
where $f_B(\cdot)$ and $T(\cdot)$ refer to the backward warping network and the trace-back function, respectively.
In implementation, we calculate the trace-back function based on the backward stereo flow. The flow represents the displacement relationship between $\tilde{I}_l^w$ and $\tilde{I}_r^w$ and is used to map the left-view mask to right-view mask.

Overall, by integrating the above two modules, our \zy{synthesis rectification} module $f_R(\cdot)$ can be expressed as below:
\begin{equation}
	M =  \left\{\begin{array}{l}
		1,\ if \ \delta^b+\delta^p > 0.9 \\
		0,\ else
		\end{array}\right.
\end{equation}
\begin{equation}
	\begin{aligned}
		\tilde{I}_r &= f_R(\tilde{I}_r^w)\\
		&= f_G((1-M)* \tilde{I}_r^w),
	\end{aligned}
\end{equation}
where $f_G(\cdot)$ represents the inpainting model that fills in the faithful content on those low confident regions.

\subsection{Loss Functions}
We train our self-rectified stereo synthesis module with pixelwise reconstruction loss imposed on both stereo warping and rectification:
\begin{equation}
	\mathcal{L}^{rec} = \sum_{channels}\frac{1}{N}\sum_{(x,y)}|I_r-\tilde{I}_r^w| + \sum_{channels}\frac{1}{N}\sum_{(x,y)}|I_r-\tilde{I}_r|
\end{equation}
where $\mathcal{L}^{rec}$ refers to the reconstruction loss that consider both warping and rectification.
The pixelwise reconstruction loss encourages the synthesized view to match the target view in pixel level.
Similarly, We train the reverse network to reconstruct the source view as follow:
\begin{equation}
	\mathcal{L}^{B} = \sum_{channels}\frac{1}{N}\sum_{(x,y)}|I_l-\tilde{I}_l^w|
\end{equation}

Besides, we leverage an adversarial loss for rectification module to enhance the generation quality of inpainted image contents for the inconsistent regions of warped images:
\begin{equation}
	\mathcal{L}^{adv} = \mathcal{L}^{adv}_R(~\tilde{I}_r),
\end{equation}
where $\mathcal{L}^{adv}$ denotes the adversarial loss of rectification module.
{Specifically, we train a generator $f_G(\cdot)$ and a discriminator $f_D(\cdot)$ as follows:
\begin{equation}
\mathcal{L}_{adv} = -{\mathbb{E}}[f_D(f_G((1-M)* \tilde{I}_r^w))],
\end{equation}
\begin{equation}
\mathcal{L}_{D} =  {\mathbb{E}}[ReLU(1+f_D(\tilde{I}_r))] + {\mathbb{E}}[ReLU(1-f_D(I_r))],
\end{equation}}

Therefore, the total loss for our self-rectified stereo synthesis module is:
\begin{equation}
	\mathcal{L} = \lambda_{rec}\mathcal{L}^{rec} + \lambda_{adv}\mathcal{L}^{adv},
\end{equation}
where $\lambda_{rec}$ and $\lambda_{adv}$ are the balancing weights for reconstruction loss and adversarial loss, respectively. 
\section{3D reconstruction}\label{sec:method_3d}
\begin{figure}[t]
	\begin{center}
		\includegraphics[width=0.99\linewidth]{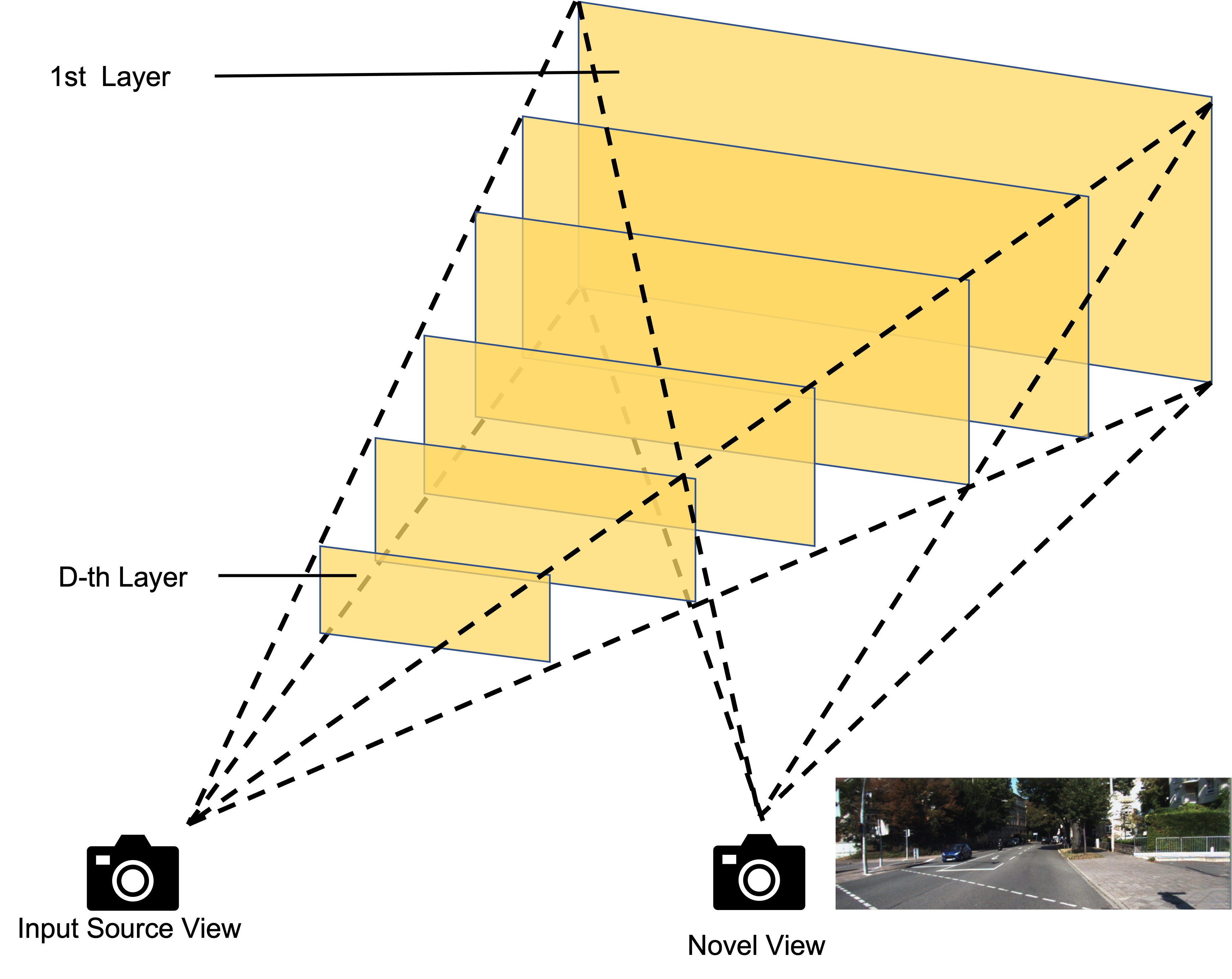}
	\end{center}
	\caption{Illustration of the MPI representation.}
	\vspace{-5mm}
	\label{fig:3d-representation}
\end{figure}

\begin{figure*}[t]
	\centering
	\setlength{\tabcolsep}{1.5pt}
	\begin{tabular}{ccc}
		\rotatebox {90}{\hspace{0.7cm}GT} &
		\includegraphics[width=.45\linewidth]{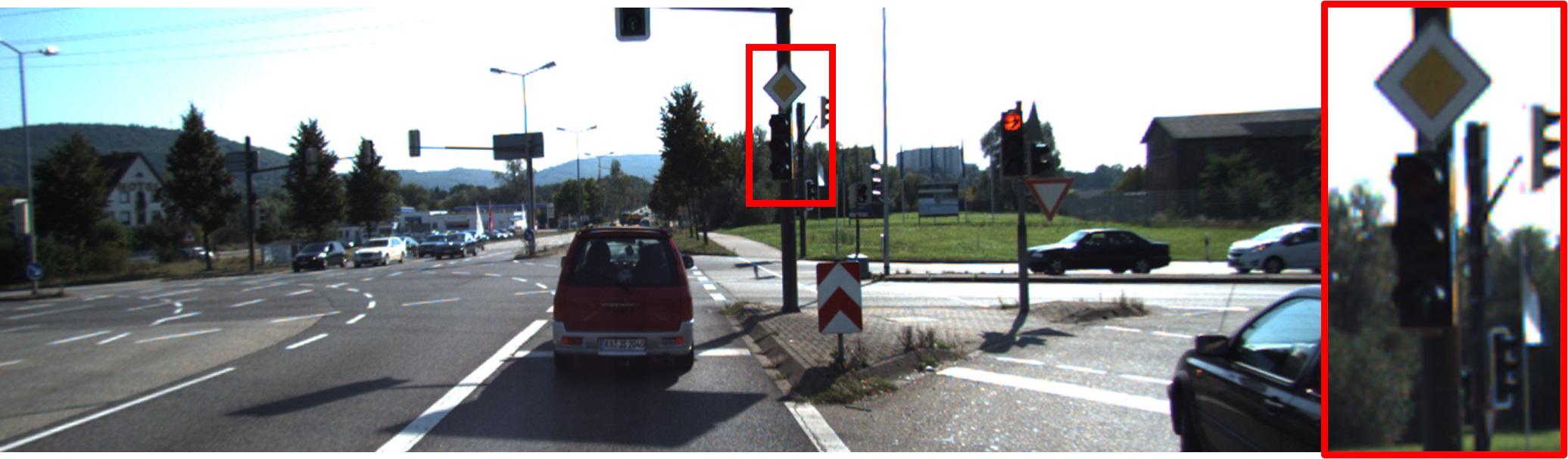} &
		\includegraphics[width=.47\linewidth]{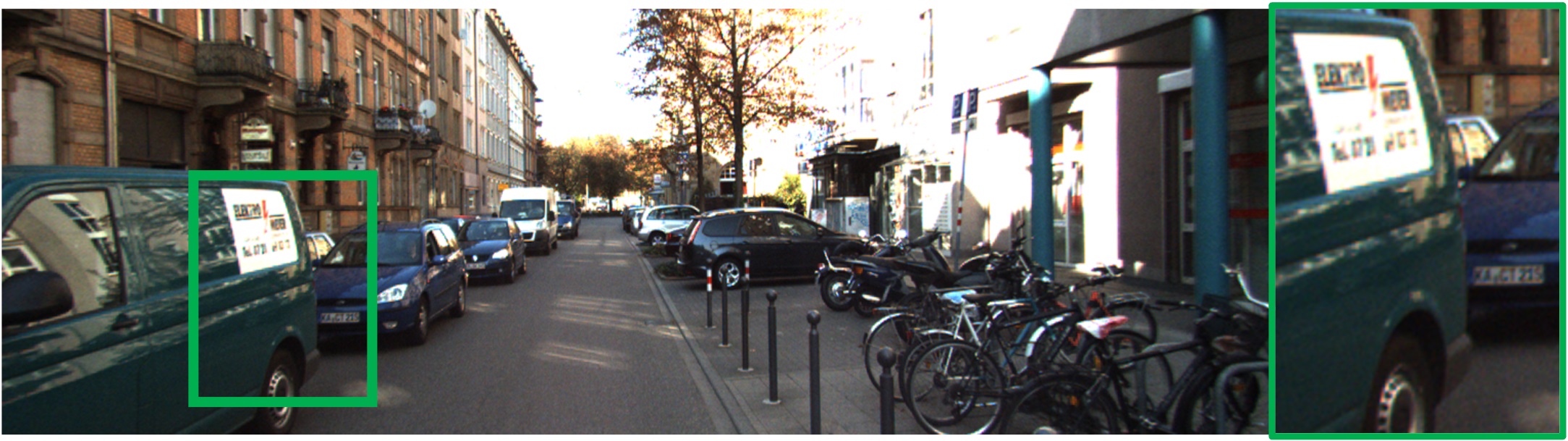} \\
		\rotatebox {90}{Godard~\etal\cite{Godard2017UnsupervisedMD} } 	 &
		\includegraphics[width=.45\linewidth]{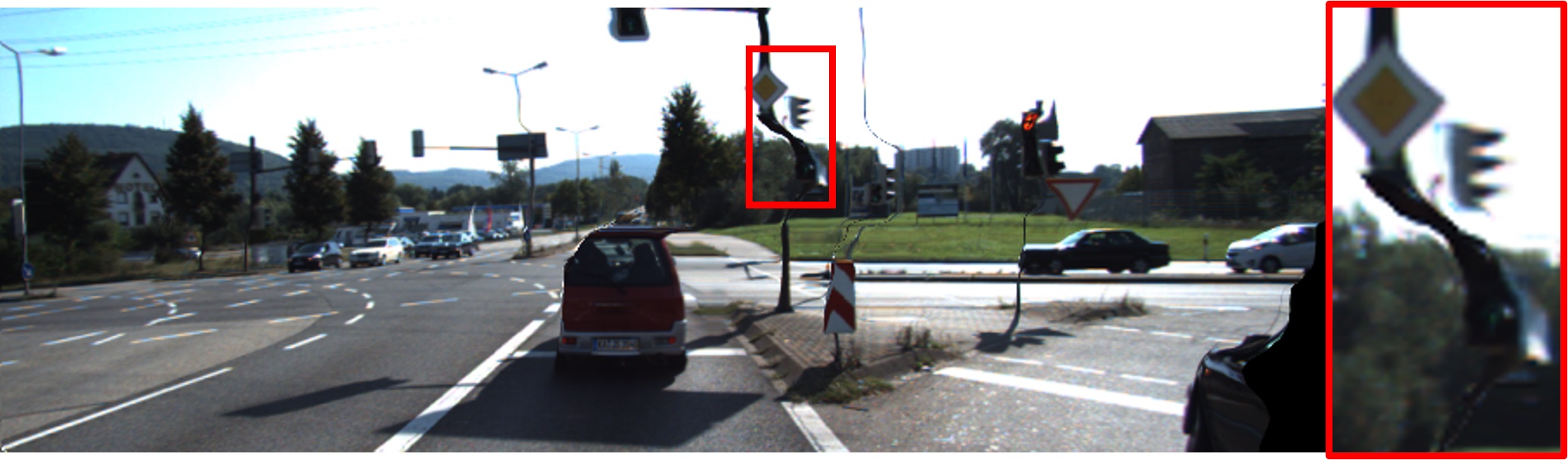} &
		\includegraphics[width=.47\linewidth]{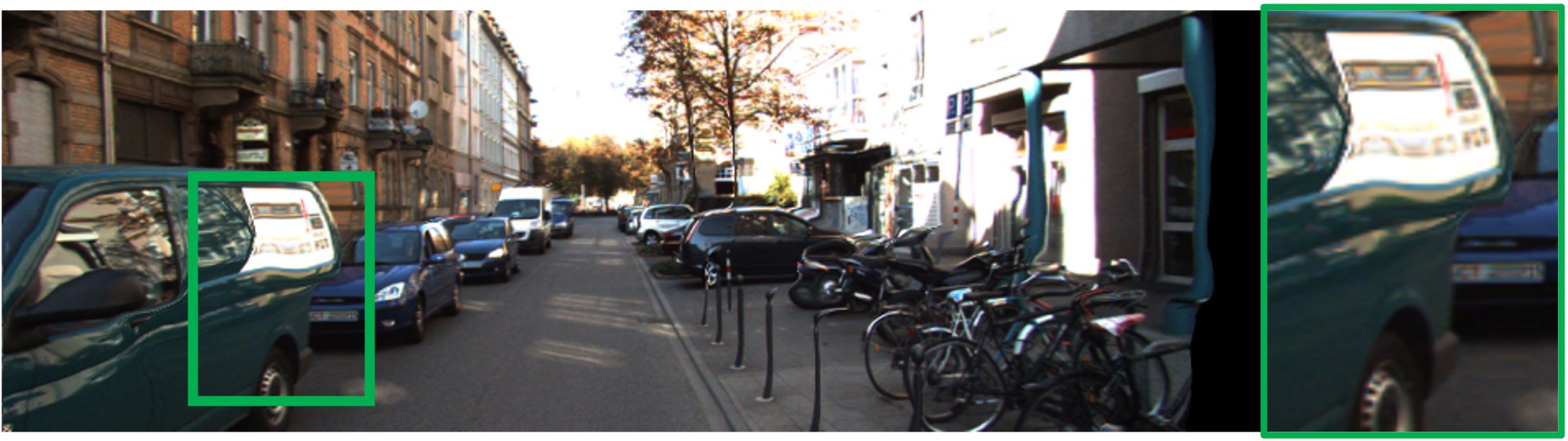}\\
		\rotatebox {90}{Xie~\etal  \cite{Xie2016Deep3DFA} }&
		\includegraphics[width=.45\linewidth]{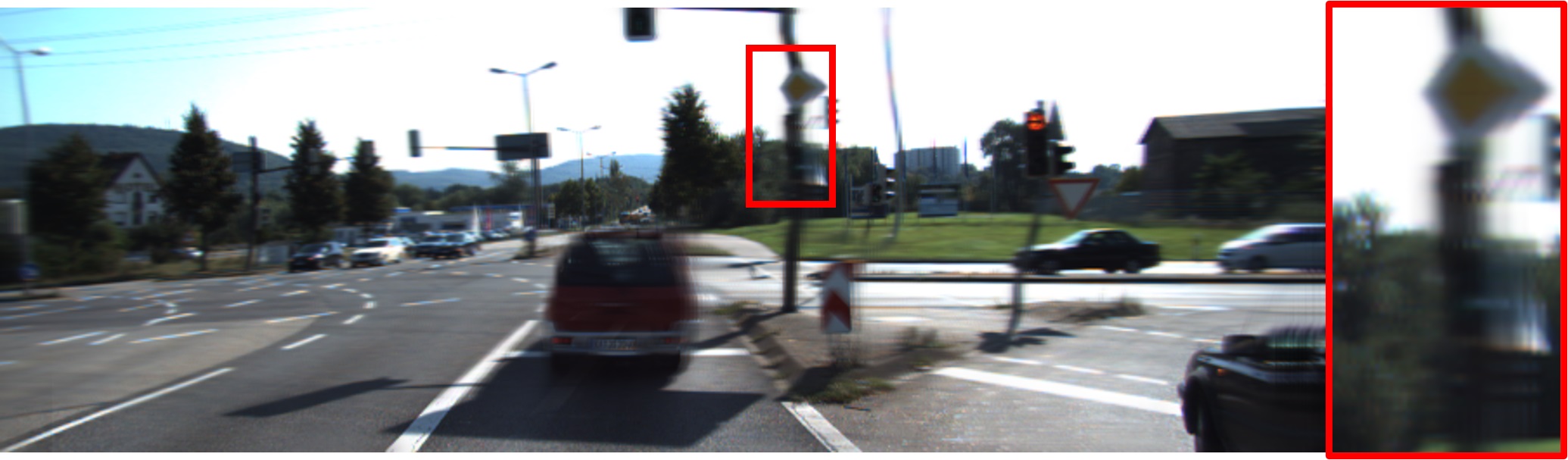} &
		\includegraphics[width=.47\linewidth]{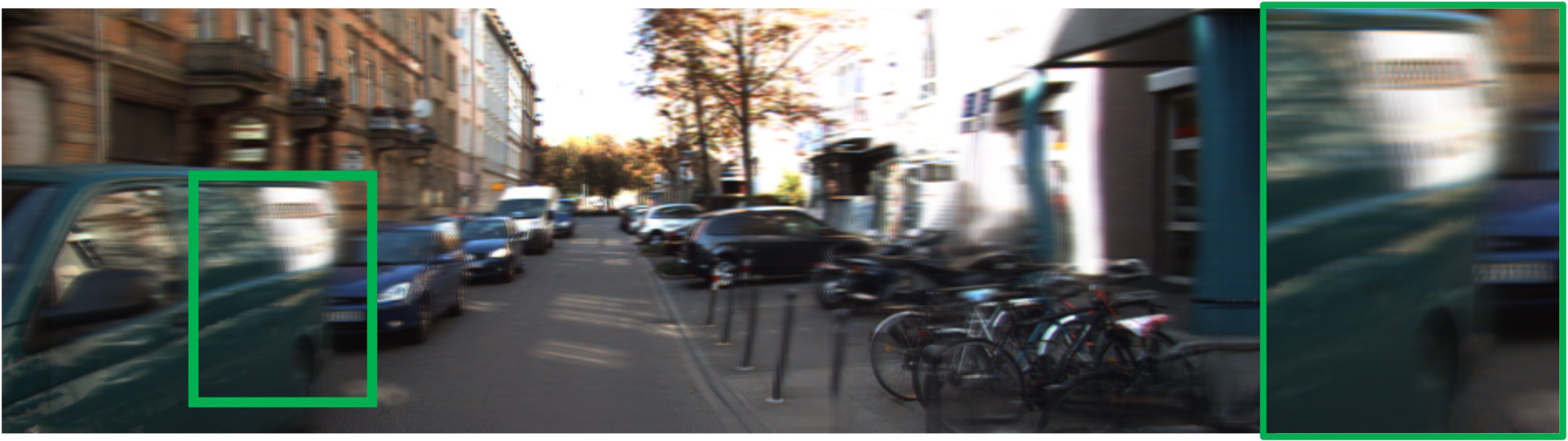}\\
	    \rotatebox {90}{Luo~\etal \cite{luo2018single} }&
		\includegraphics[width=.45\linewidth]{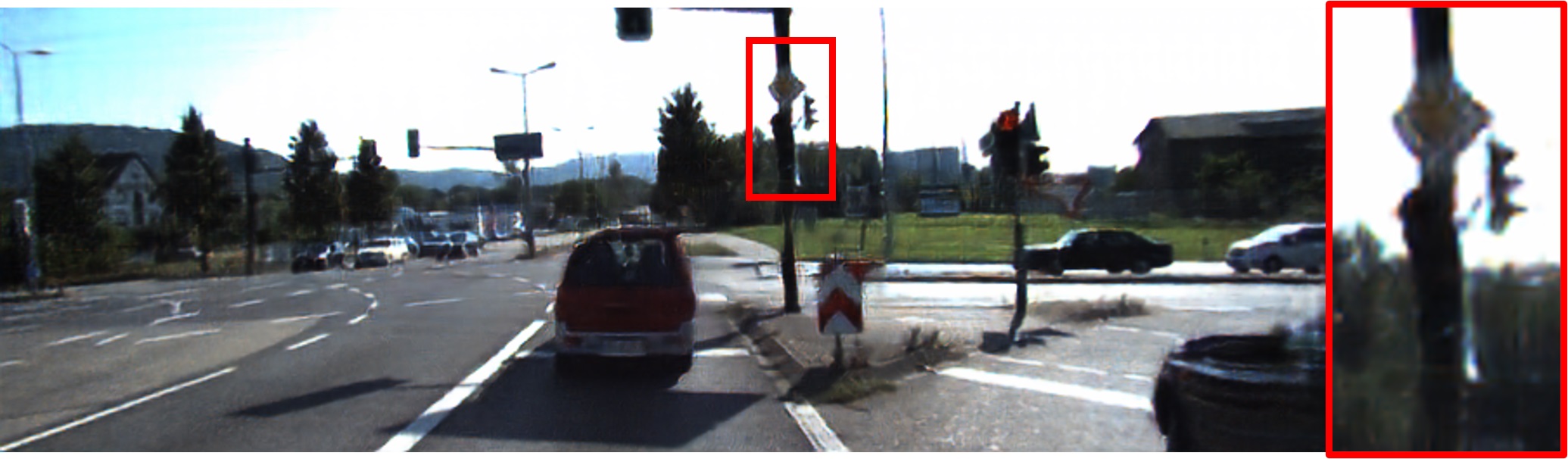} &
		\includegraphics[width=.47\linewidth]{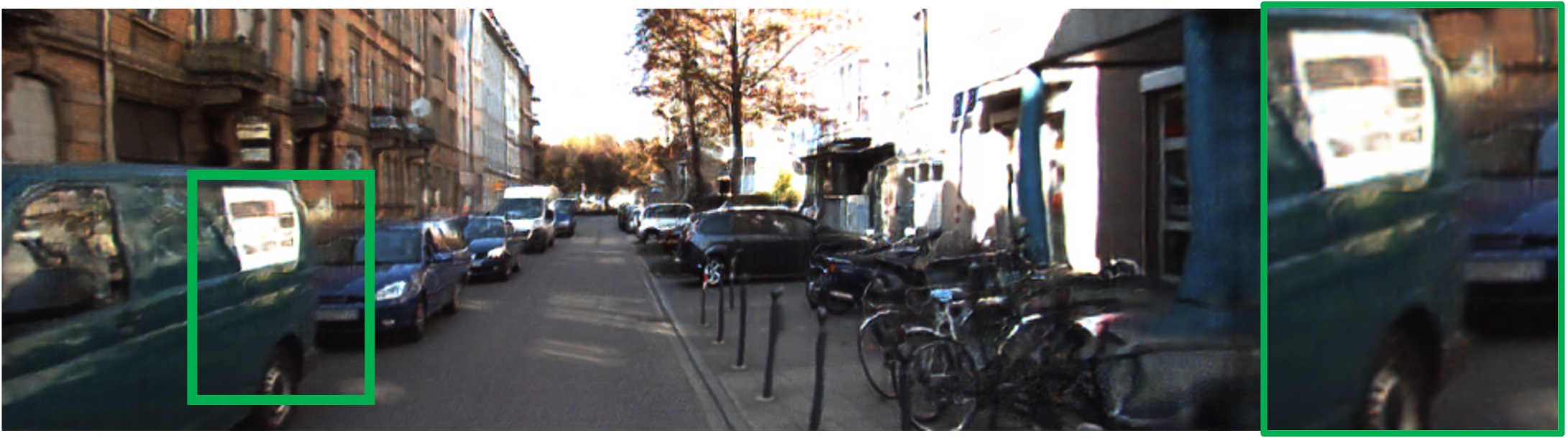}\\
		\rotatebox {90}{\zy{Gonzalez~\etal \cite{gonzalezbello2020forget}}}&
		\includegraphics[width=.45\linewidth]{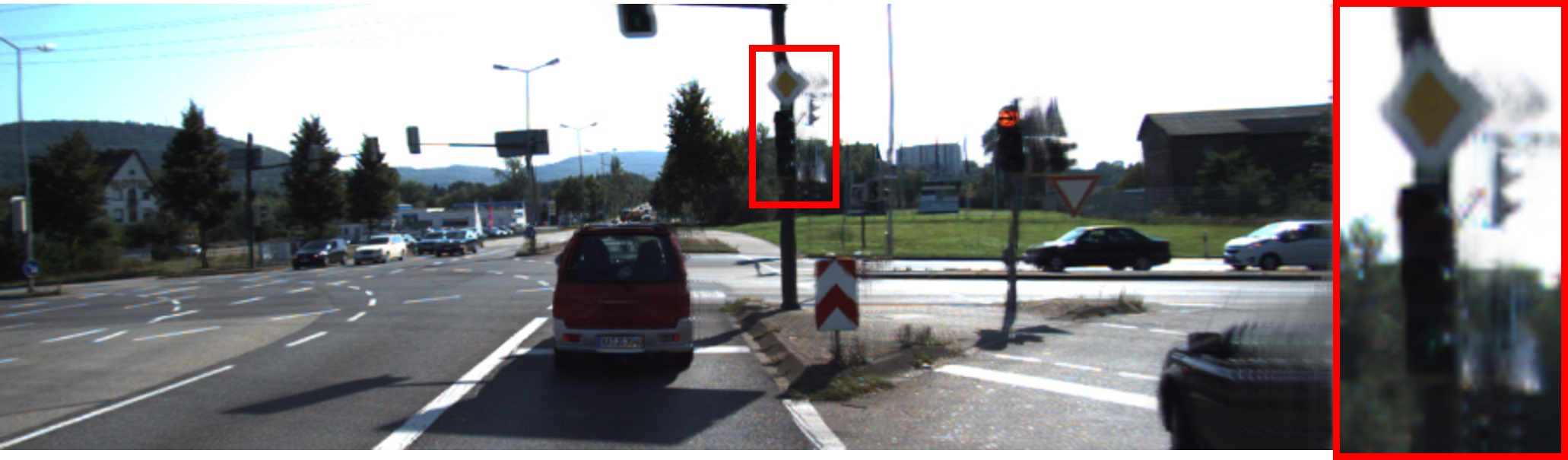}  &
		\includegraphics[width=.47\linewidth]{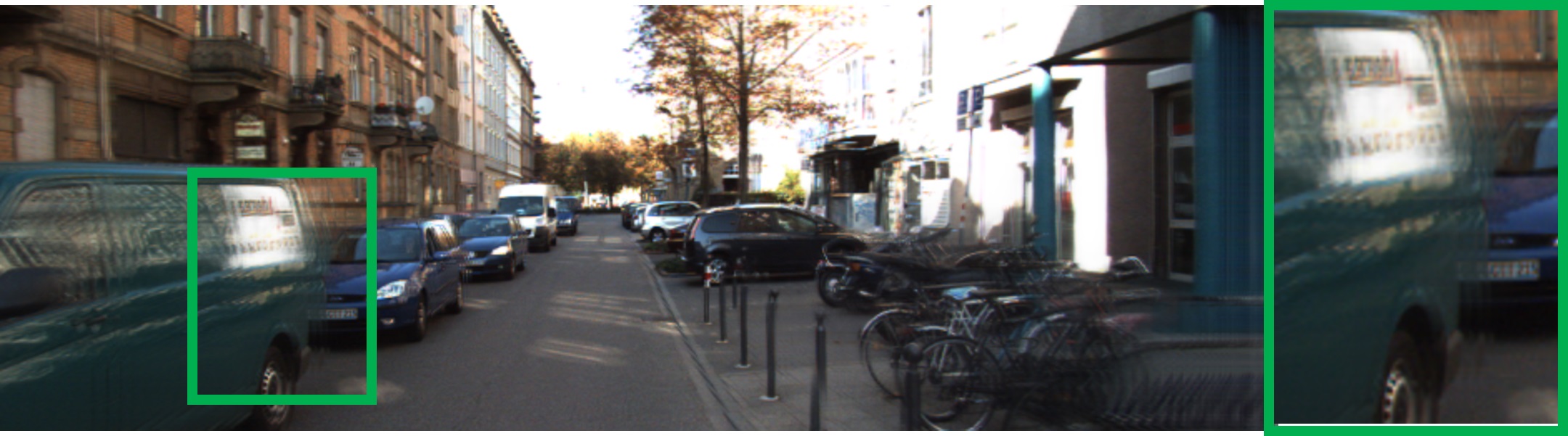} \\
		\rotatebox {90}{\zy{Gonzalez~\etal \cite{Gonzalez_2021_CVPR}}}&
		\includegraphics[width=.45\linewidth]{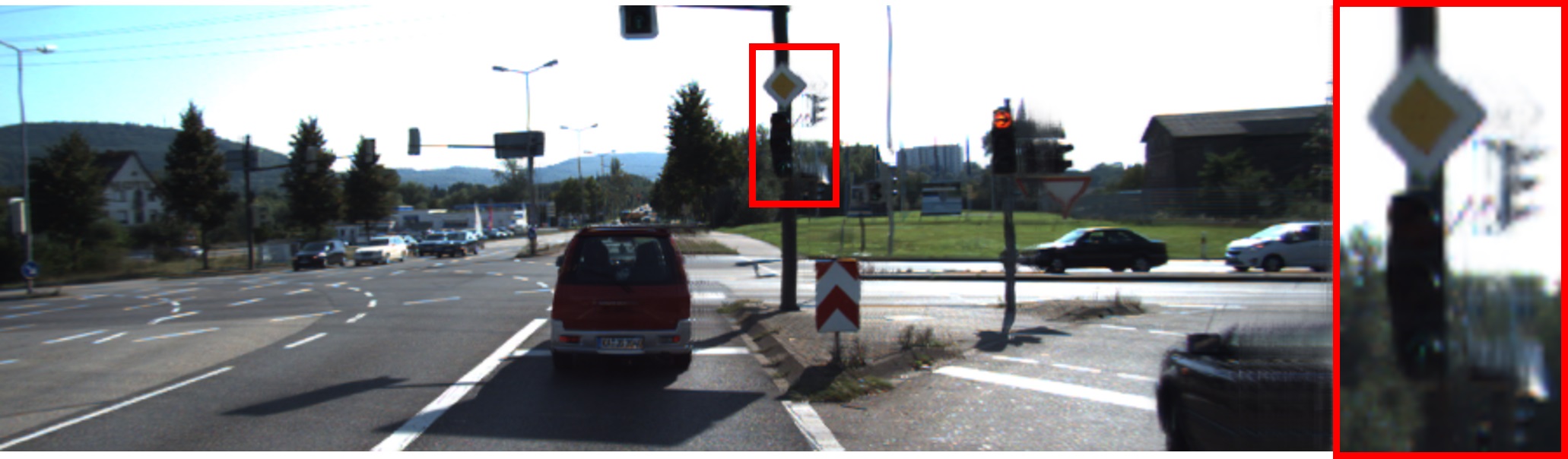}  &
		\includegraphics[width=.47\linewidth]{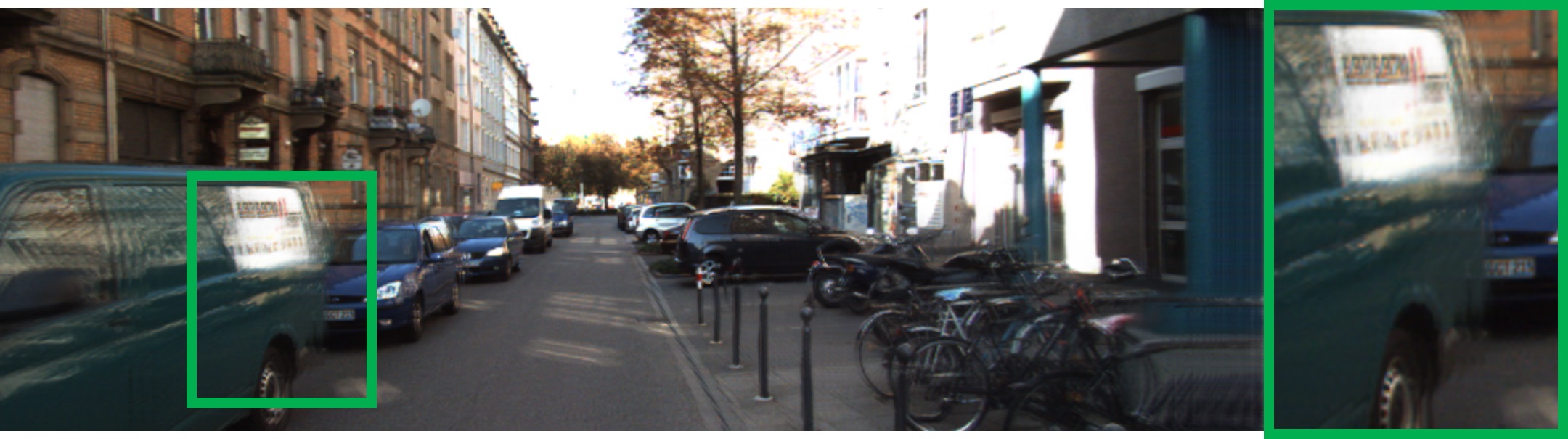} \\
		\rotatebox {90}{\hspace{0.6cm}Ours}&
		\includegraphics[width=.45\linewidth]{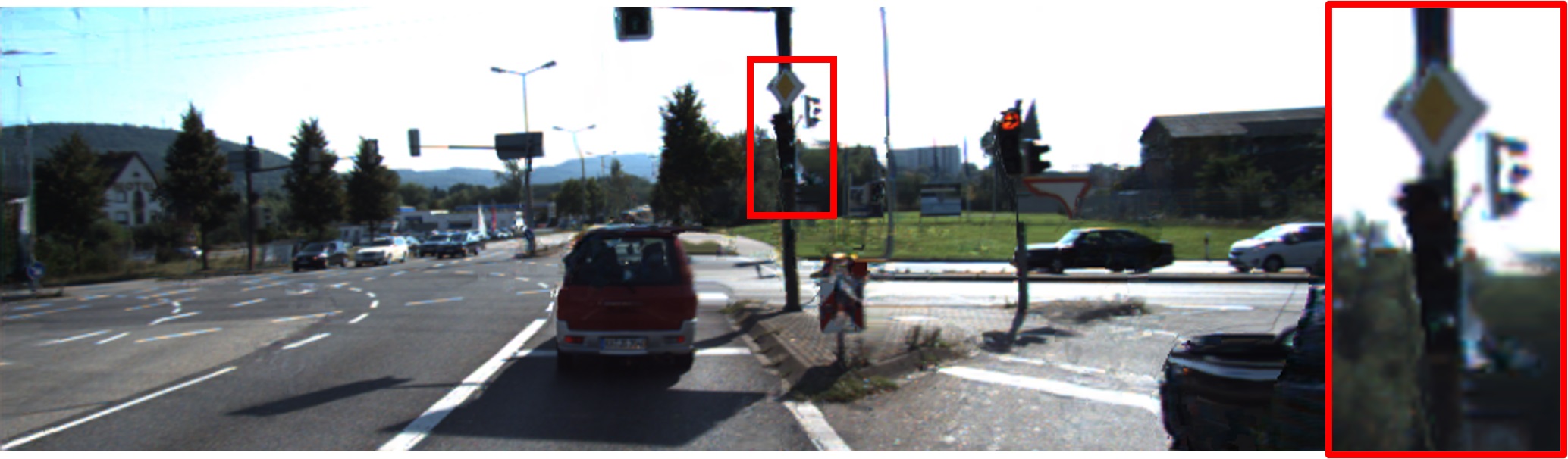}  &
		\includegraphics[width=.47\linewidth]{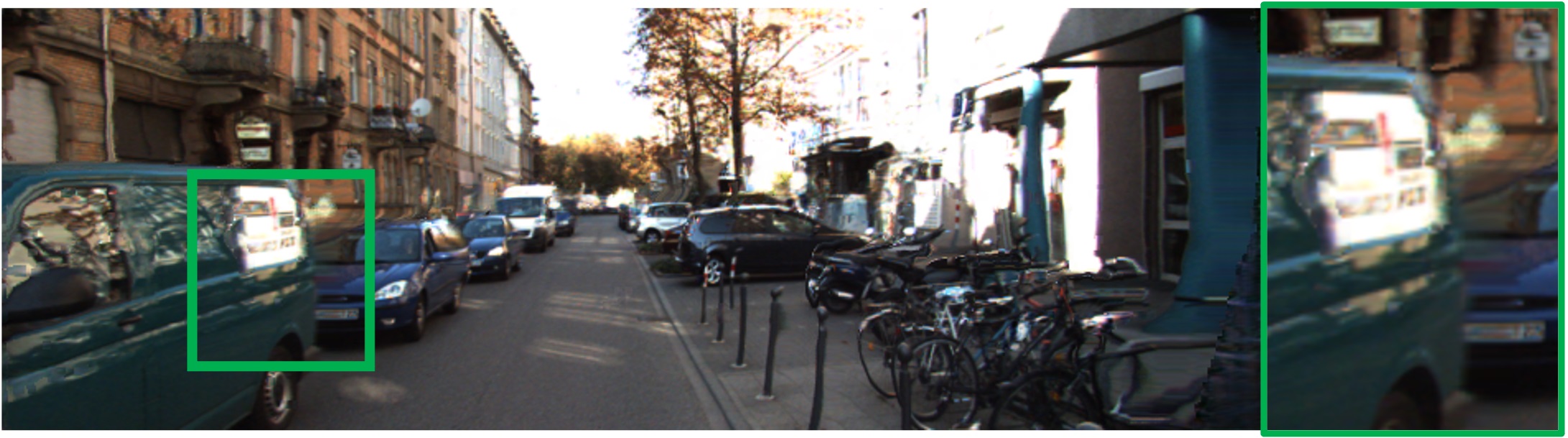} \\
		
	\end{tabular}
	\caption{Qualitative comparison of stereo synthesis against state-of-the-art methods.}
	\vspace{-2mm}
	\label{fig:stereo-synthesis-table-visual}
\end{figure*}

In this section, we describe our 3D representation as shown in Figure \ref{fig:3d-representation} that used in 3D reconstruction.
Our approach is allowed to work with arbitrary 3D representations. Here we choose one of the recently used MPI as the representation for 3D reconstruction.

In specific, MPI is a set of $D$ fronto-parallel RGB-$\alpha$ image planes with respect to a reference camera view $v_{ref} $, in which the plane $i$ is at the fixed depth $d_i$ and encodes the color and transparency $(c_i, \alpha_i)$ of that depth, \ie, $\{(c_1, \alpha_1), (c_D, \alpha_D)\}$. To render a novel view from an MPI, we perform the inverse homography operation on each plane, in order to transform them to the target camera viewpoint. Then, we apply back-to-front alpha composition on the transformed planes for the target view. Note that, inverse homography and alpha composition are both differentiable, so that they are easy to incorporate into our end-to-end learning framework.

Concretely, we adopt the 3D representation following \cite{Srinivasan2019PushingTB}. Unlike previous MPI prediction methods that use a 2D network structure, our 3D representation prediction network is based on a 3D convolutional encoder-decoder network structure with skip connections and a dilated convolution \cite{Yu2016MultiScaleCA}, in favor to properly fuse depth information of plane and spatial information.
\zy{During the 3D reconstruction process, we first convert the pseudo-stereo visual pair into a plane-sweep-volume tensor (PSV) with the shape of [6, H, W, D], where 6 is the color channel of the stereo inputs, H and W denote the height and width, and D represents the number of image planes.
Then, we employ the 3D representation prediction network to obtain the MPI. The 3D convolutional network takes the PSV tensor as input and output a tensor with the shape of [4, H, W, D], \ie, the predicted MPI, where 4 denotes the RGB-$\alpha$ channels.} We are able to generate novel views based on the resulting MPI representation through rendering techniques such as homography warping and integration of the planes.

\section{Experiments}\label{sec:exp}
In this section, we first present qualitative and quantitative comparisons on stereo synthesis to validate our \zy{synthesis rectification} strategies.
Moreover, we carry out experiments with the MPI representation for demonstrating the effectiveness of stereo prior on single-view view synthesis task.
In addition, ablation experiment is set up to further analyze the functions of two different rectification strategies.
{Besides, we also demonstrate the effectiveness of our method using the LDI representation.}

\zy{As for the implementation details,
we follow the design of Flownet~\cite{ilg2017flownet} for the warping network and reverse network. Specifically, the network is an auto-encoder with skip connections between encoder and decoder. It takes the left view as the input and predicts a flow corresponding to the right view, which is further used to generate the right view with the backward warping.
To facilitates our arbitrary mask inpainting, we adopt the inpainting model proposed by \cite{yu2019free} as our self-rectified network. It is composed of two auto-encoders: the first network generates the coarse inpainted results with gated convolution, and the second network utilizes the contextual attention to synthesize the refined results based on the coarse results.}
While training the networks, we set the weighting factors $\lambda_{rec}=1$ and $\lambda_{adv}=0.1$ and adopt Adam optimizer \cite{kingma2014adam} with a learning rate of 0.0001. The magnitude-based pruning percentage $p$ is 0.5. Besides, the 3D reconstruction network is a 3D CNN predicts an initial MPI~\cite{Srinivasan2019PushingTB}.
During the experiment with MPI, we set the number of image planes $D=32$.
Following \cite{Tucker2020SingleViewVS,li2021mine}, we train our method with 20 of the city image sequences from the KITTI raw dataset \cite{Geiger2013VisionMR} at resolution $768\times256$ and test on the other 4 city image sequences.
\zy{The above networks are trained from scratch.}
During testing, in lines with the settings of the previous methods~\cite{Tucker2020SingleViewVS,li2021mine}, we perform the qualitative metrics at resolution $384\times128$ and crop all sides of the images by 5\% of the height and width respectively.

\subsection{Stereo Synthesis Comparison}

\begin{table}[tb]
	\centering
	\caption{Quantitative comparison of stereo synthesis on KITTI dataset.}
	\setlength{\tabcolsep}{0.1cm}{
	\begin{tabular}{c|c|c|c}
		\toprule
	Method	& PSNR $\uparrow$ & SSIM $\uparrow$ & LPIPS $\downarrow$\\ \midrule\midrule
		Xie~\etal \cite{Xie2016Deep3DFA} & 19.9 & 0.780 & 0.175\\
		Godard~\etal \cite{Godard2017UnsupervisedMD} & 19.5 & 0.768 & 0.169\\
		Luo~\etal \cite{luo2018single} & 20.3 & 0.791 & 0.156\\
		Gonzalez~\etal \cite{gonzalezbello2020forget} & 21.6 & 0.807 & 0.134\\
		Gonzalez~\etal \cite{Gonzalez_2021_CVPR} & 21.9 & 0.816 & 0.128\\
		Ours & \textbf{22.1} & \textbf{0.824}  & \textbf{0.124}\\  \midrule
		Baseline &20.1 & 0.783 & 0.161\\
		w/ Bidirectional matching & 21.3 &0.808 & 0.143\\
		w/ Pruning & 21.0 & 0.800  & 0.140 \\ \midrule
		$p=10\%$ &21.5 & 0.812 & 0.131\\
		$p=30\%$ &21.8 & 0.817 & 0.129\\
		$p=70\%$ &22.0 & 0.820 & 0.126\\
		$p=90\%$ &21.6 & 0.814 & 0.132\\
		\bottomrule
	\end{tabular}
	}
	\label{fig:stereo-synthesis-table}
\end{table}

We compare our stereo synthesis results with five previous works \cite{Godard2017UnsupervisedMD,Xie2016Deep3DFA,luo2018single,gonzalezbello2020forget,Gonzalez_2021_CVPR}. We add two strategies (network pruning and bidirectional matching) to the stereo synthesis process to ensure that better results can be generated. The quantitative results in terms of PSNR, SSIM, and LPIPS perceptual similarity~\cite{zhang2018unreasonable} are shown in Table \ref{fig:stereo-synthesis-table}. We can see that the best results are achieved by our method among these methods.
The previous methods aim at predicting a better disparity-like map. Although they perform well on visible regions, they are inefficient on handling the disoccluded regions.
Thanks to the two rectification strategies, our method is able to identify the error regions caused by the first stage and inpaint it with reasonable contents.
Specifically, the pruning-based rectification detects the intractable area of warping while the bidirectional matching rectification finds out the inconsistent pixels between the stereo views.
On the other hands, the exemplar synthesis results are shown in Figure \ref{fig:stereo-synthesis-table-visual}, and we can see that there are some artifacts in their results. In particular, their synthesized stereo either produces blurring artifacts, or contains severe distortion.
In Figure \ref{fig:stereo-synthesis-table-visual}, the poles generated by Xie~\etal \cite{Xie2016Deep3DFA} and Godard~\etal \cite{Godard2017UnsupervisedMD} (highlighted in the red rectangles) are distorted, while the ones generated by our method are very close to the ground-truth image. \zy{Although the pole in the synthesized results produced by Luo~\etal~\cite{luo2018single} and Gonzalez~\etal~\cite{gonzalezbello2020forget} are straight, their image details are not well preserved.
Also, the shape of traffic lights in Gonzalez~\etal~\cite{Gonzalez_2021_CVPR} is blurry.}
Moreover, the contour of the van in the green rectangles generated by previous methods is blurry or distorted, while our method can produce faithful result.

\subsection{MPI-based Novel View Synthesis Comparison}

\begin{table}[t]
	\centering
	\caption{Quantitative comparison of novel view synthesis on KITTI dataset. The ideal case represents the novel view synthesis using real stereo image pairs and represents the upper limit of our algorithm. \zy{Ours-2D denotes our model using 2D CNN as 3D reconstruction network.}}
	\begin{tabular}{c|c|c|c}
		\toprule
	Method	& PSNR $\uparrow$ & SSIM $\uparrow$ & LPIPS $\downarrow$ \\ \midrule\midrule
				Ideal Cases & 30.7 & 0.984 & 0.014 \\ \midrule
		Tucker~\etal \cite{Tucker2020SingleViewVS} & 19.5 & 0.733 & - \\
		MINE \cite{li2021mine} & 21.4 & 0.822 & 0.112\\
		\zy{Ours-2D} & \zy{22.0} & \zy{0.829} & \zy{0.112}\\
		Ours & \textbf{22.6} & \textbf{0.840} & \textbf{0.110}\\
		\bottomrule
	\end{tabular}
	\label{fig:View-Synthesis-on-KITTI}
	\vspace{-2mm}
\end{table}

In the following, we compare with latest single-view MPI reconstruction methods~\cite{li2021mine,Tucker2020SingleViewVS}.
For a fair comparison, we follow the same evaluation protocol, \ie, training at resolution $768\times256$ and testing at resolution $384\times128$, with 32 image planes of MPI.
For quantitative comparison, we adopt the metrics PSNR, SSIM, and LPIPS perceptual similarity~\cite{zhang2018unreasonable} to evaluate the reconstruction results.
The quantitative results are shown in Table \ref{fig:View-Synthesis-on-KITTI}. As observed,
the ideal cases are based on ground-truth stereo pairs, and thus it reflects the upperbound of our approach and demonstrates the effectiveness of our decoupling strategy.
Although there is gap between our method and the ideal case, our method still achieves superior performance against the previous methods.
The results indicate that explicitly imposing the stereo prior knowledge with pseudo-stereo pairs can mitigate the ambiguity of single-view MPI reconstruction.
Besides, the self-supervised stereo warping network also strengthens the detail preservation in the synthesized results.

\begin{figure}[t]
	\centering
	\setlength{\tabcolsep}{1pt}
	\begin{tabular}{ccc}
		\rotatebox{90}{\hspace{0.35cm}GT} &
		\includegraphics[width=.47\linewidth]{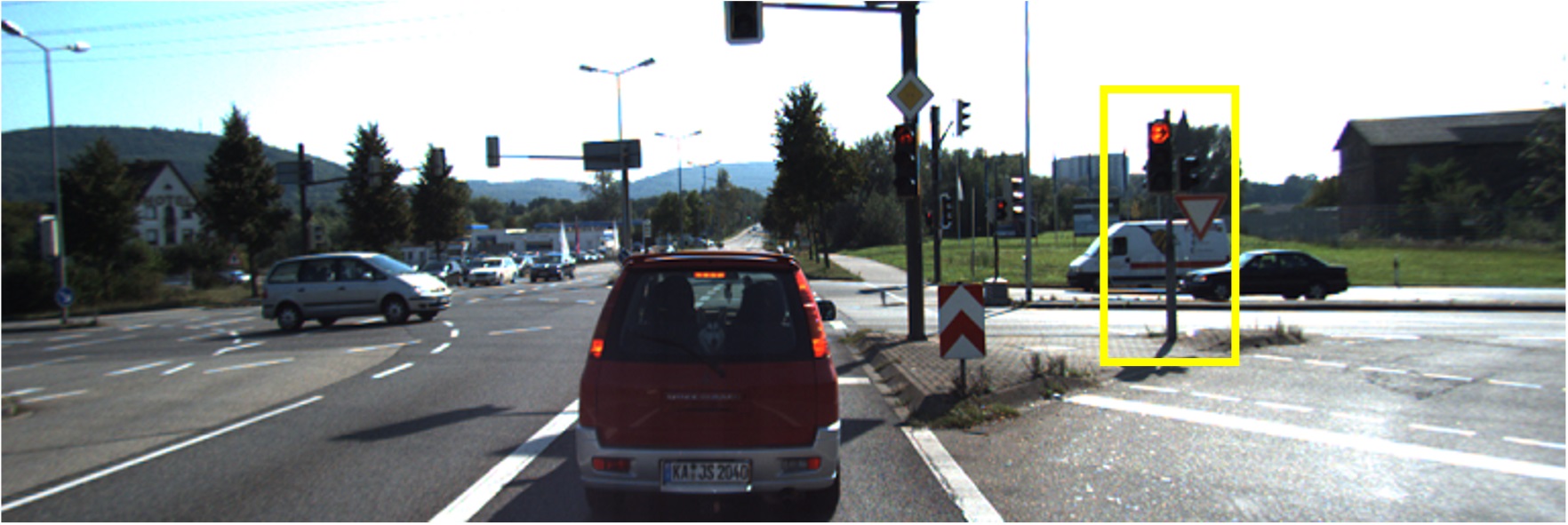} &
		\includegraphics[width=.47\linewidth]{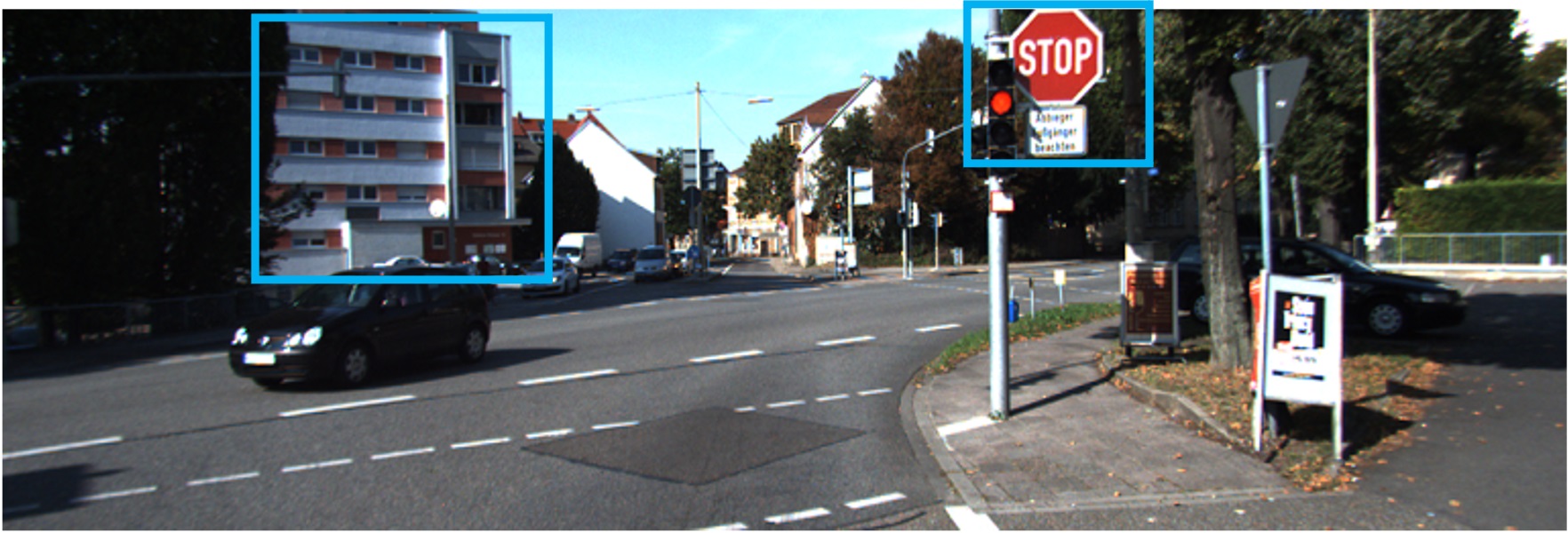} \\
		\rotatebox{90}{\hspace{0.38cm}\cite{Tucker2020SingleViewVS}} &
		\includegraphics[width=.47\linewidth]{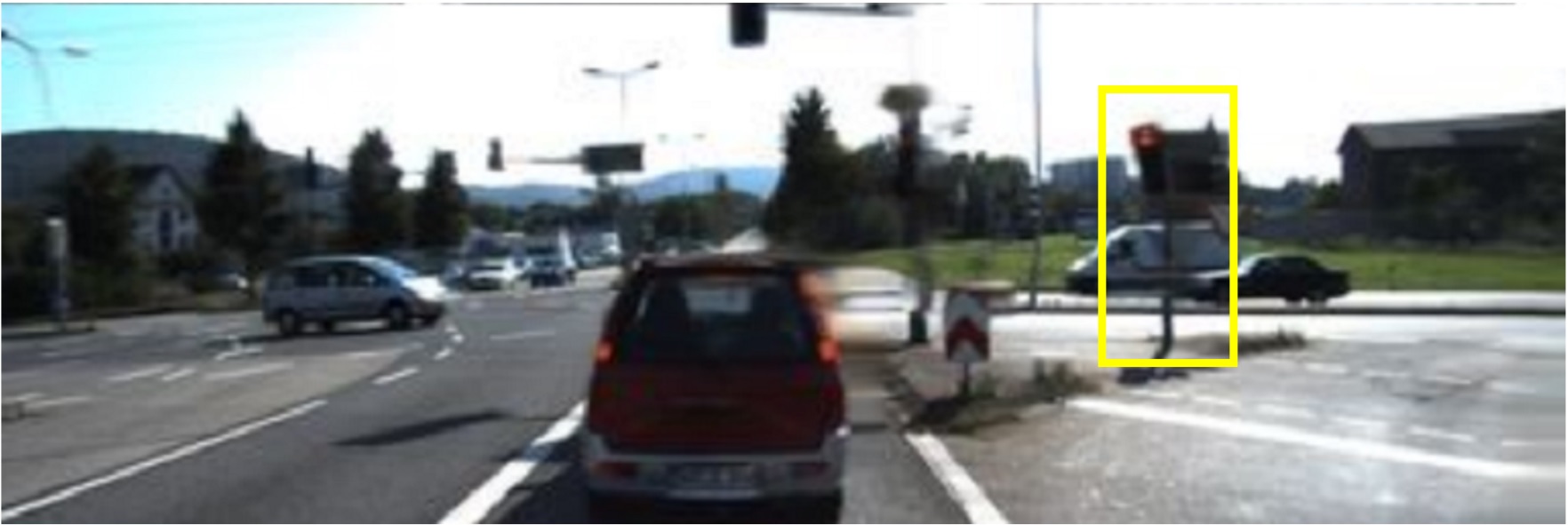} &
		\includegraphics[width=.47\linewidth]{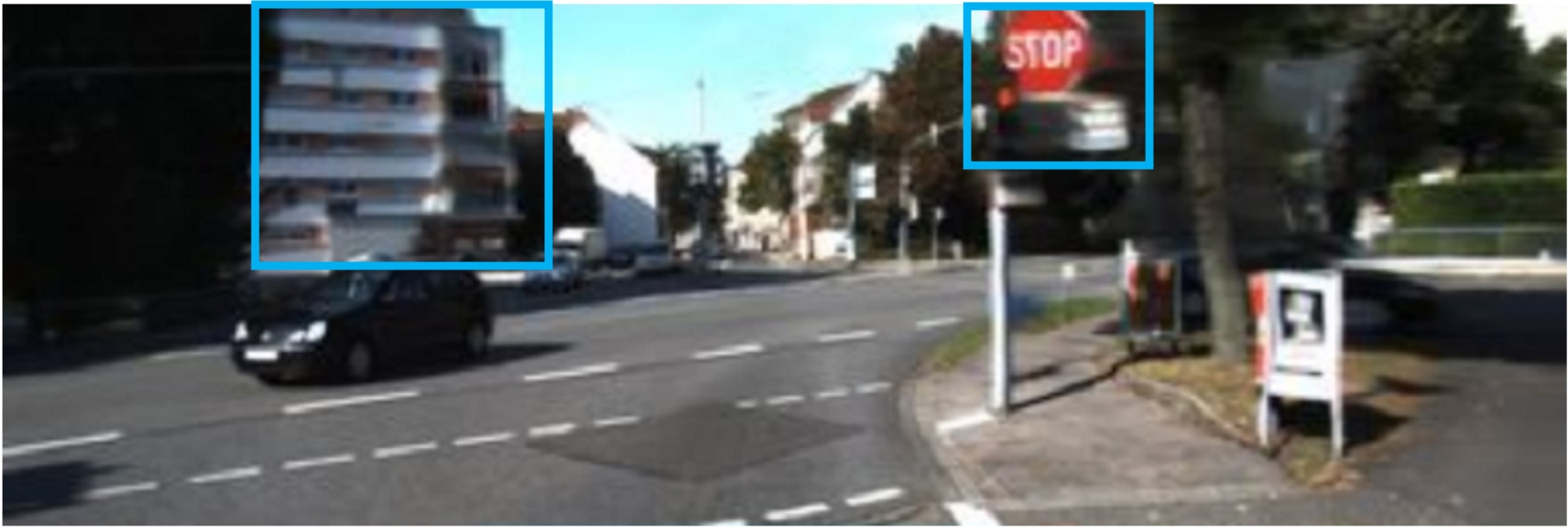} \\
		\rotatebox {90}{\hspace{0.38cm}\cite{li2021mine}} &
		\includegraphics[width=.47\linewidth]{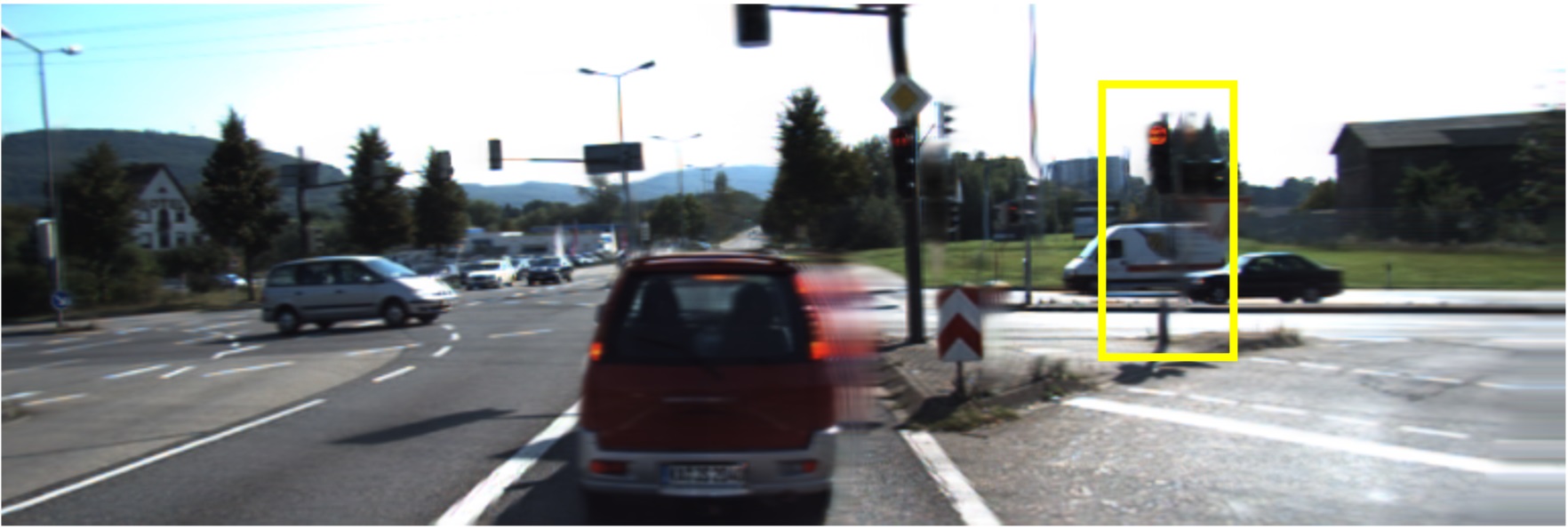} &
		\includegraphics[width=.47\linewidth]{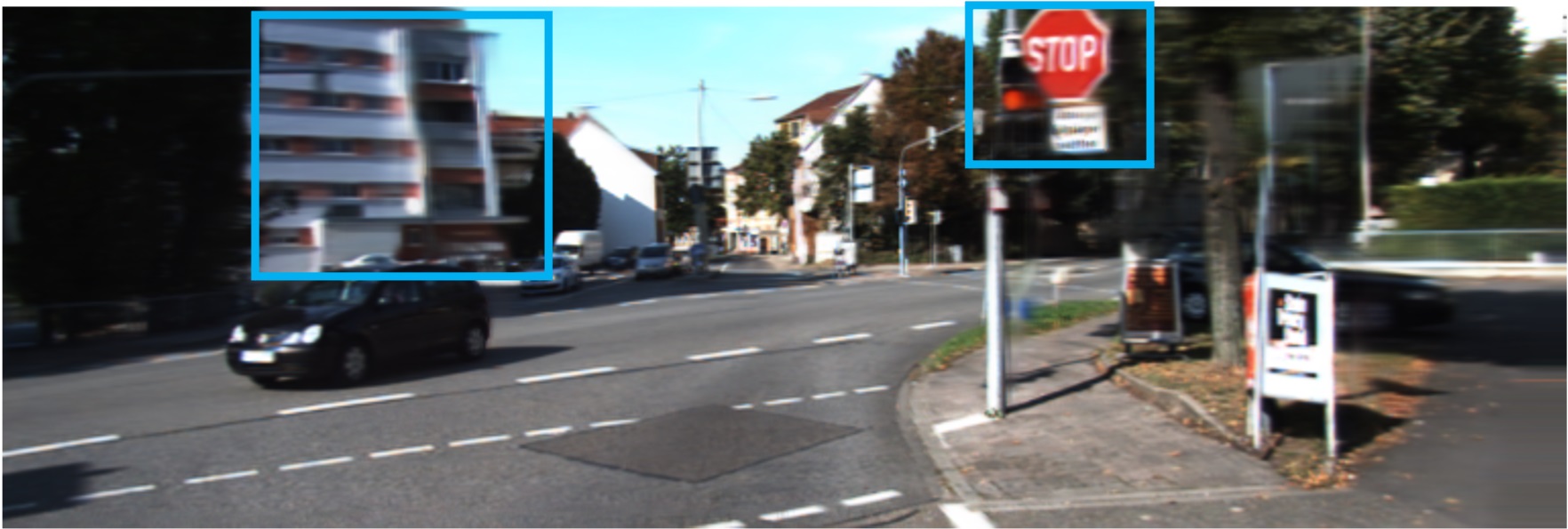} \\
		\rotatebox{90}{\hspace{0.35cm}Ours} &
		\includegraphics[width=.47\linewidth]{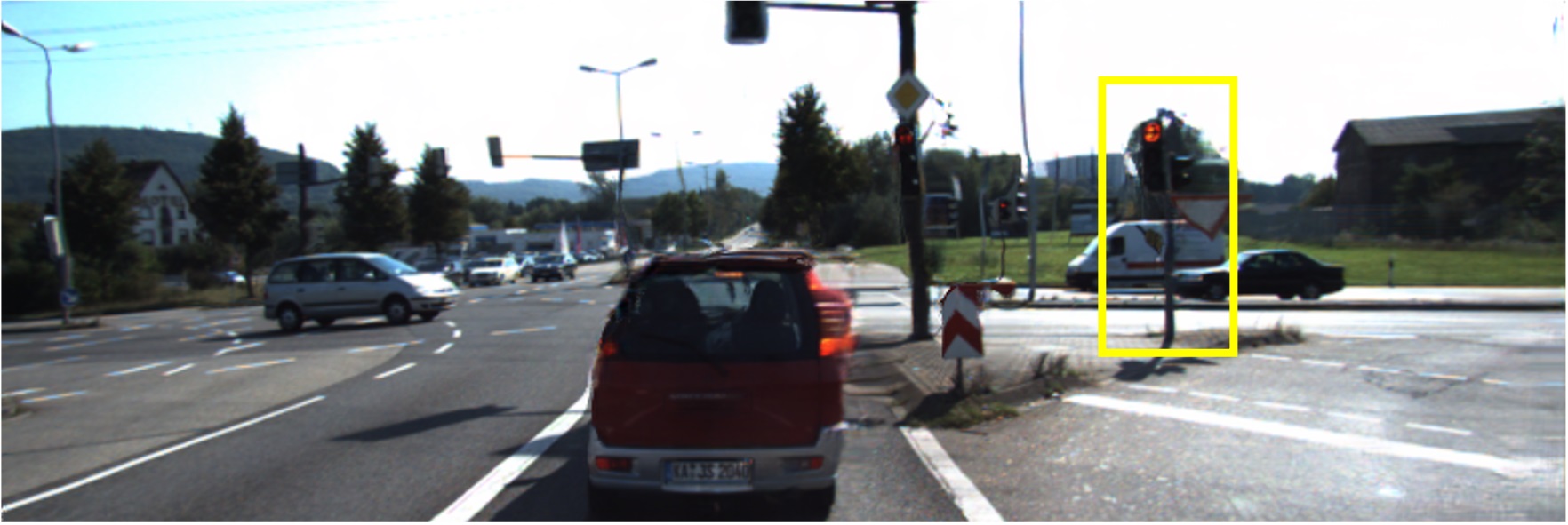} &
		\includegraphics[width=.47\linewidth]{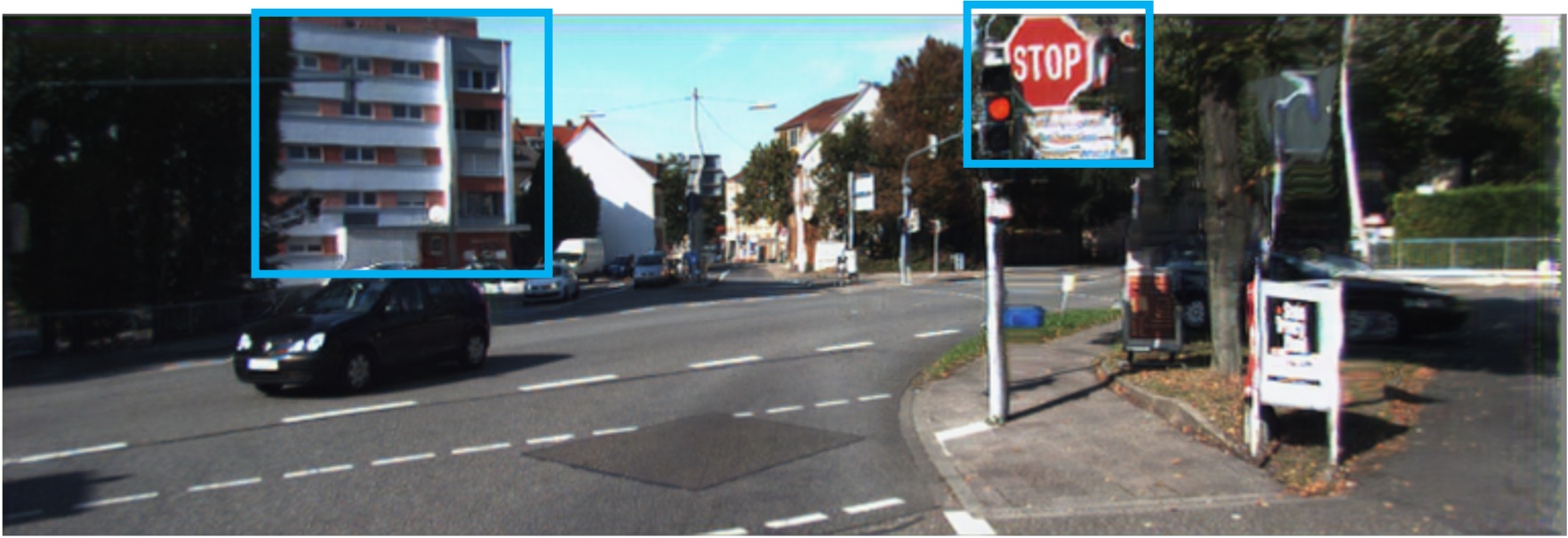} \\
	\end{tabular}
	\caption{Qualitative comparison of novel view synthesis against other single-view MPI learning methods. These results are not cherry-picked but the same images used in~\cite{Tucker2020SingleViewVS}. (Better viewed in digital version with zoom function.)}
	\label{fig:View-Synthesis-on-KITTI-visual}
\end{figure}

Furthermore, we provide the qualitative comparison in Figure \ref{fig:View-Synthesis-on-KITTI-visual}. Note that, as the source code of \cite{Tucker2020SingleViewVS} is not publicly available, we directly compare with the representative results collected from their original paper, rather than cherry-picked ones.
As demonstrated, the self-rectified stereo synthesis module reduces the ambiguity and complexity in single-view view synthesis.
From the figure we can see that our pseudo-stereo method produce more clear and realistic texture while less structure distortions. Particularly, around the disoccluded region,
ours is able to inpaint more clear content with the aid of stereo prior. In contrast, both Tucker~\etal \cite{Tucker2020SingleViewVS} and MINE~\cite{li2021mine} generate blurry results on these regions.

\subsection{Ablation Study}

In this section, we delve into the two \zy{synthesis rectification} strategies, pruning-based and bidirectional matching rectification.
We consider three variants and conduct experiments to analyze the design of our method:
\zy{1) baseline model, in which the warped image is simply adopted as the pseudo right-view image;
2) w/ pruning; 3) w/ bidirectional matching; 4) our complete model.
}

As shown in Table \ref{fig:stereo-synthesis-table},
both pruning-based rectification and bidirectional matching rectification gains improvements from the baseline,
which suggests that rectification is in favor of improving the image quality of pseudo view.
Comparing these two rectification strategies, we find that bidirectional matching rectification performs slightly better than pruned-based rectification, as it not only infers the distorted structures, but also identifies the inconsistent region.
Yet, combining these strategies together yields the optimal results, especially the obviously improved SSIM and LPIPS, since both strategies can compensate each other.

\begin{figure}[t]
	\centering
	\setlength{\tabcolsep}{1.3pt}
	\begin{tabular}{cc}
		\multicolumn{2}{c}{\includegraphics[width=.49\linewidth]{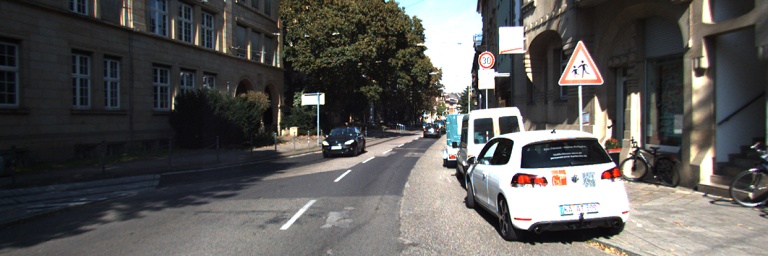}}\\
		\multicolumn{2}{c}{GT} \\
		\includegraphics[width=.49\linewidth]{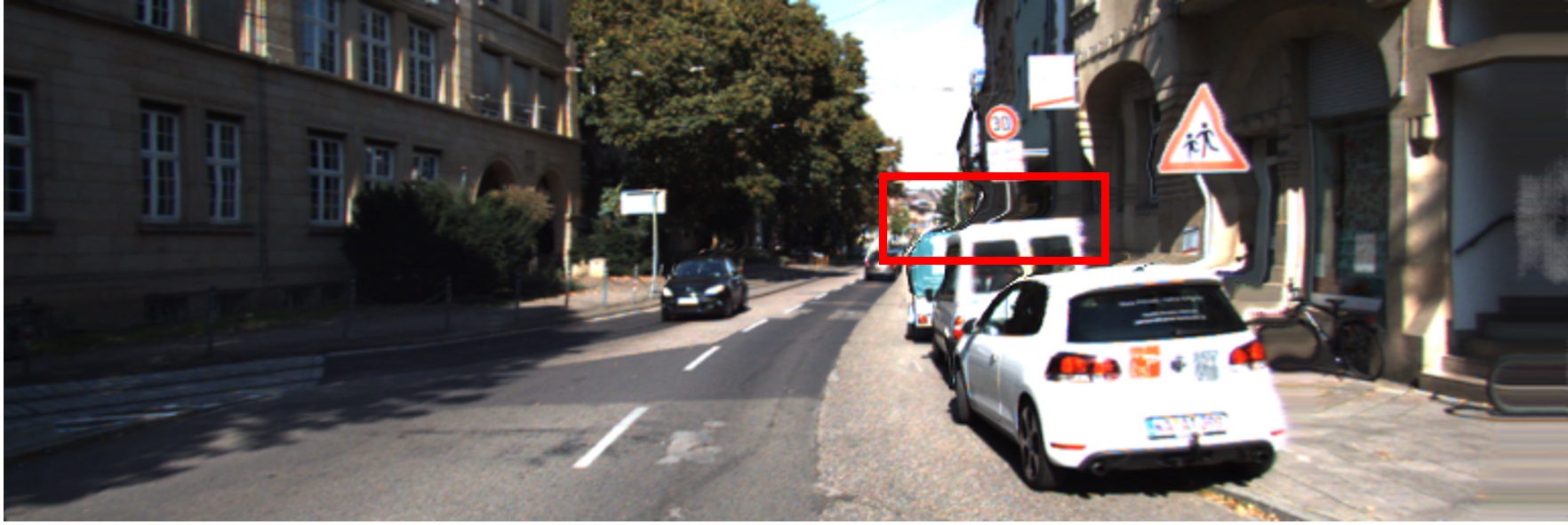} &
		\includegraphics[width=.49\linewidth]{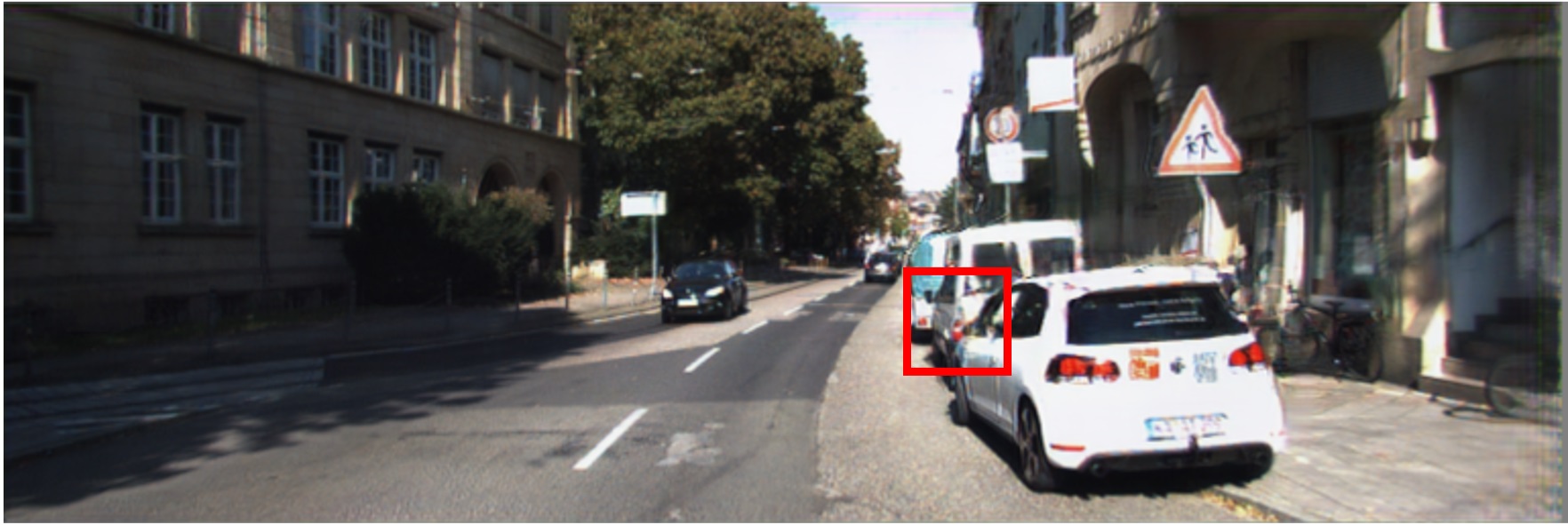} \\
		Baseline & w/ Pruning \\
		\includegraphics[width=.49\linewidth]{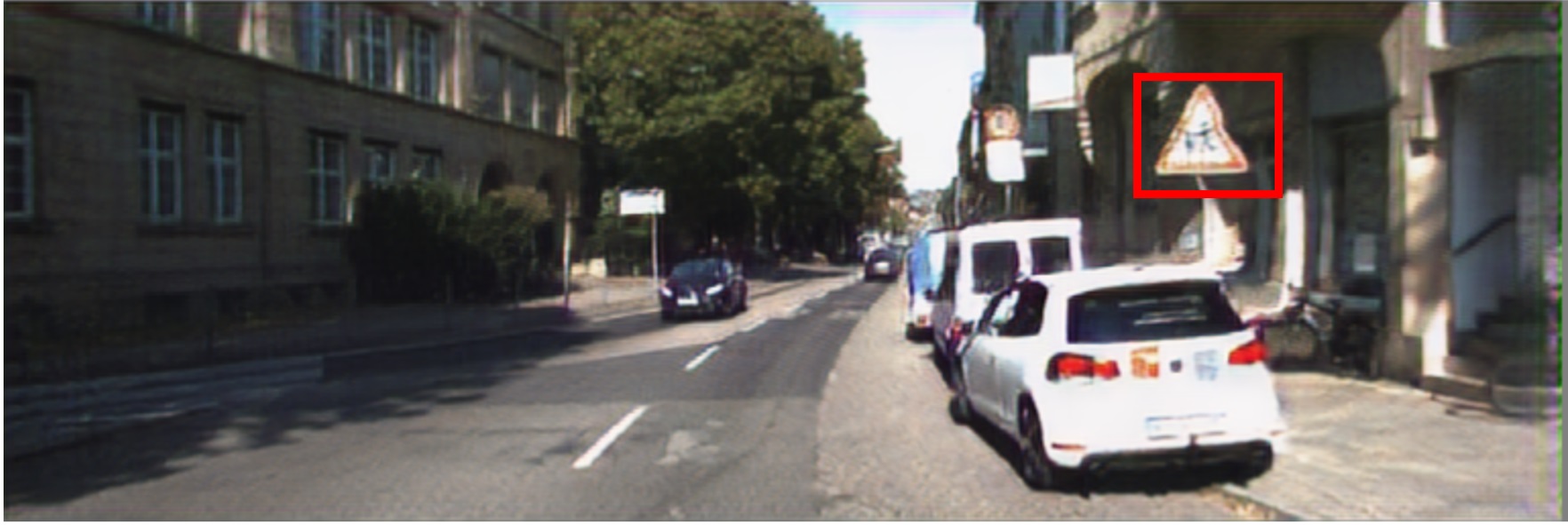} &
		\includegraphics[width=.49\linewidth]{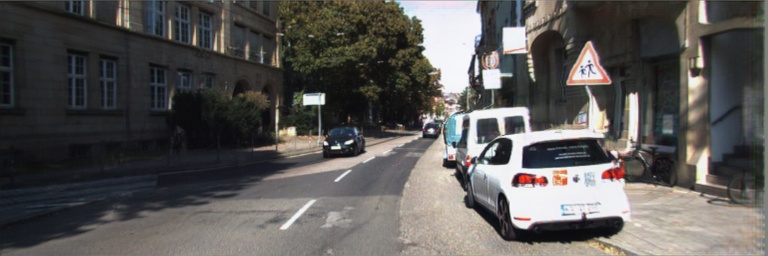} \\
		w/ Bidirectional  & Complete model
	\end{tabular}
	\vspace{-1mm}
	\caption{Quantitative results of ablation studies on rectification modules.}
	\vspace{-3mm}
	\label{fig:ablation-visual}
\end{figure}

We also provide qualitative results for ablation studies, as shown in Figure \ref{fig:ablation-visual}.
As illustrated, although the baseline method preserves clear details, it contains obvious artifacts due to incorrect warping\zy{, such as the crooked pole.}
Pruning-based rectification can mitigate the warping artifacts by inpainting the hard-to-train region with authentic contents,
and the bidirectional matching rectification further improve this results by rectifying the inconsistent regions.
\zy{However, there are noises and blurs in the synthesized results. As highlighted by the red rectangles, the content near the front of car is unclear and the sign is also blurry.}
Our complete model take advantages of all strategies and produce more realistic stereo view.

In addition, we carry out experiments to analyze the influence of pruning percentage $p$,
resulting in 5 variants: $p = \{10\%, 30\%, 50\%, 70\%, 90\%\}$. Note that, $p=50\%$ is the complete model.
Quantitative results are shown in Table \ref{fig:stereo-synthesis-table}. When we increase the pruning percentage from 10\% to 50\%, more hard-to-train regions will be detected and rectified, leading to a better image quality.
However, the image quality becomes lower if we keep increasing the pruning percentage from 50\% to 90\%, indicating excessive pruning impairs the performance of warping network on the easy-to-warp regions.
Therefore, we set $p=50\%$ in our complete model to achieve best image quality.

\zy{To verify the efficacy of 3D convolution in 3D reconstruction, we replace the 3D convolution with 2D convolution in the reconstruction network. As shown in Table \ref{fig:View-Synthesis-on-KITTI}, the pseudo-stereo images help our 2D convolutional network reaching a better performance compared to prior models. However, our method is able to gain best results using 3D convolution. It indicates that 3D convolution is in favor of leveraging depth and spatial information jointly.
} 
{
\subsection{LDI-based Novel View Synthesis Comparison}
Our approach is allowed to work with arbitrary 3D representations. To verify this, other than the MPI representation, we compare with Tulsiani~\etal\cite{Tulsiani2018Layerstructured3S} based on Layered-Depth-Image (LDI) representation.
We follow the original architecture of Tulsiani~\etal\cite{Tulsiani2018Layerstructured3S} that predicts a two-layers LDI using CNN.
However, we modify the network input from a single view input to stereo inputs for taking advantage of the stereo prior.
The other experiment settings are the same as Tulsiani~\etal\cite{Tulsiani2018Layerstructured3S}.
}

\begin{table}[t]
	\centering
	\caption{{Quantitative comparison of novel view synthesis on different representations and datasets.}}
	\setlength{\tabcolsep}{0.1cm}{
	\begin{tabular}{c|c|c|c|c}
		\toprule
	   Dataset     &Method	& PSNR $\uparrow$ & SSIM $\uparrow$ & LPIPS $\downarrow$ \\ \midrule\midrule
	  \multirow{2}*{KITTI}      &Tulsiani~\etal\cite{Tulsiani2018Layerstructured3S} & 16.5 & 0.572 & 0.179\\
		~          &Ours & \textbf{18.6} & \textbf{0.633} & \textbf{0.150}\\ \midrule
		\multirow{2}*{Scene Flow} &MINE \cite{li2021mine} & 15.7 & 0.568 & 0.244\\
		 ~         &Ours & \textbf{17.0} & \textbf{0.640} & \textbf{0.194}\\
		\bottomrule
	\end{tabular}
	}
	\label{fig:LDI-comparision}
\end{table}

\begin{figure}[t]
	\centering
	\setlength{\tabcolsep}{1.5pt}
	\begin{tabular}{ccc}
		\rotatebox {90}{\hspace{0.35cm}GT} &
		\includegraphics[width=.47\linewidth]{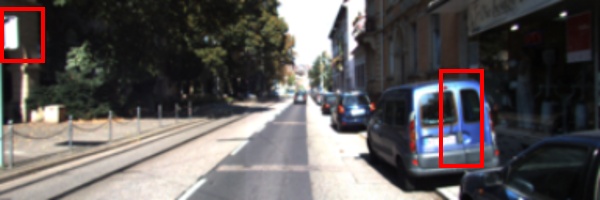} &
		\includegraphics[width=.47\linewidth]{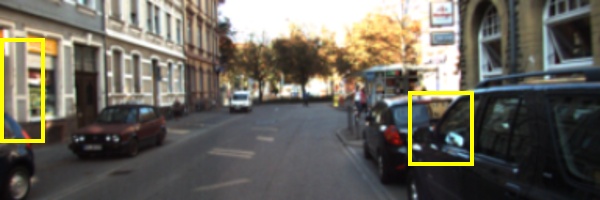} \\
		\rotatebox {90}{\hspace{0.35cm}\cite{Tulsiani2018Layerstructured3S}} &
		\includegraphics[width=.47\linewidth]{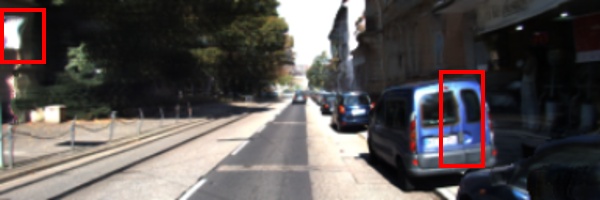} &
		\includegraphics[width=.47\linewidth]{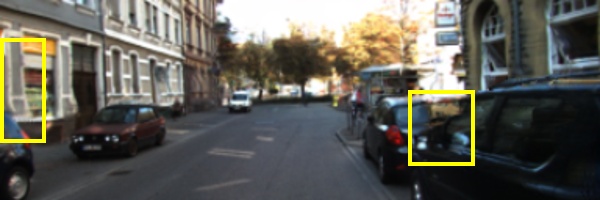} \\
		\rotatebox {90}{\hspace{0.35cm}Ours} &
		\includegraphics[width=.47\linewidth]{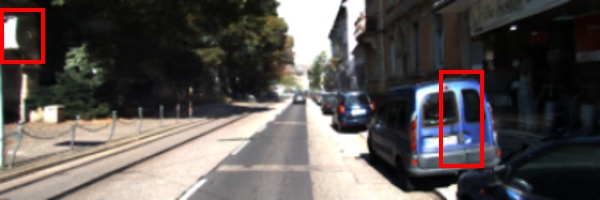} &
		\includegraphics[width=.47\linewidth]{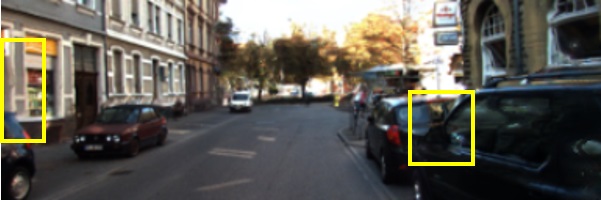} \\
	\end{tabular}
	\caption{{Qualitative comparison of LDI-based novel view synthesis against Tulsiani~\etal\cite{Tulsiani2018Layerstructured3S}. (Better viewed in digital version with zoom function.)}}
	\vspace{-2mm}
	\label{fig:View-Synthesis-on-KITTI-visual-ldi}
\end{figure}

\begin{figure}[t]
	\centering
	\setlength{\tabcolsep}{1.5pt}
	\begin{tabular}{ccc}
		\rotatebox {90}{\hspace{0.35cm}GT} &
		\includegraphics[width=.47\linewidth]{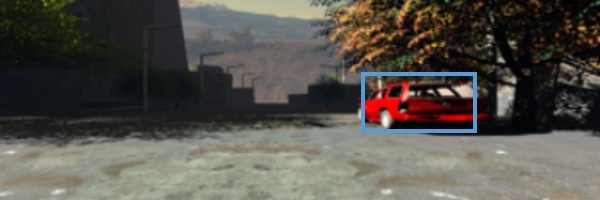} &
		\includegraphics[width=.47\linewidth]{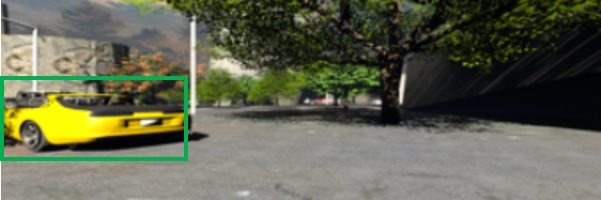} \\
		\rotatebox {90}{\hspace{0.35cm}\cite{li2021mine}} &
		\includegraphics[width=.47\linewidth]{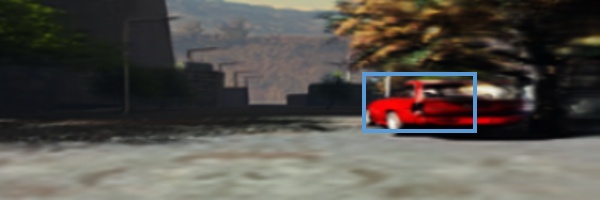} &
		\includegraphics[width=.47\linewidth]{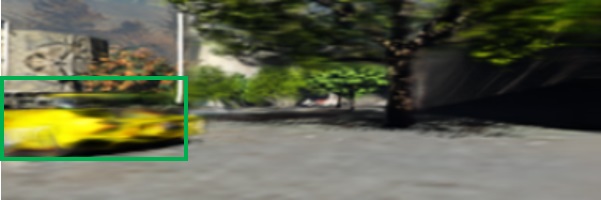} \\
		\rotatebox {90}{\hspace{0.35cm}Ours} &
		\includegraphics[width=.47\linewidth]{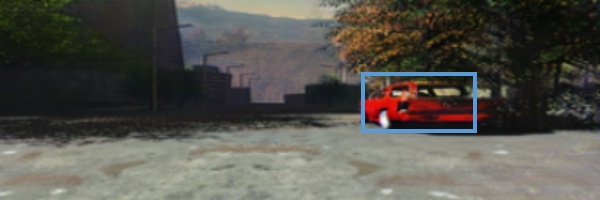} &
		\includegraphics[width=.47\linewidth]{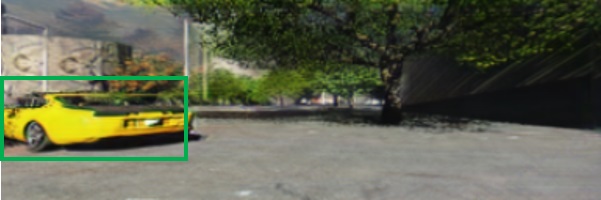} \\
	\end{tabular}
	\caption{{Qualitative comparison of novel view synthesis on the Scene Flow dataset~\cite{mayer2016large}. (Better viewed in digital version with zoom function.)}}
	\vspace{-2mm}
	\label{fig:View-Synthesis-on-Driving-visual}
\end{figure}

{We present the quantitative comparison results on the image sequences of KITTI raw dataset in Table~\ref{fig:LDI-comparision}.
We can see that our method surpasses Tulsiani~\etal~on both PSNR and SSIM metrics. It shows that our pseudo-stereo plays an positive role in the process of LDI prediction by introducing stereo prior.
Furthermore, it provides evidence that our proposed method is general enough to be applied to the other 3D representations.
}

{
Qualitative results are shown in Figure~\ref{fig:View-Synthesis-on-KITTI-visual-ldi}.
As can be seen, there are more structure distortions in the results of Tulsiani~\etal\cite{Tulsiani2018Layerstructured3S}.
While our method effectively leverage the stereo prior to alleviate the ambiguity from the single view.
As a result, we are able to generate novel views with more plausible geometry and texture.
} 
{
\subsection{Evaluation on the Scene Flow Dataset}
We conduct an experiment on the Scene Flow dataset~\cite{mayer2016large} to further validate the generalization of the proposed method.
The dataset consists of more than 39000 stereo frames, rendered from various synthetic sequences.
Specifically, we sample the training data and testing data randomly from the driving sequences of Scene Flow in $384\times128$ pixel resolution.
}

{
As shown in Table \ref{fig:LDI-comparision}, our proposed method also outperforms MINE~\cite{li2021mine} in this dataset, further demonstrating our effectiveness. Figure~\ref{fig:View-Synthesis-on-Driving-visual} presents the qualitative comparison results on this dataset.
We can see that our method is capable to synthesize sharper edges and clearer textures with the guidance of pseudo stereo pair.
However, MINE~\cite{li2021mine} fails to generate faithful details as the lack of auxiliary information.
} 
\zy{
\subsection{Evaluation on the Flowers Light Fields}
In order to verify the capability of our method on images beyond stereo test pairs, we also carry out an experiment on the Flowers light fields dataset~\cite{srinivasan2017learning}.
The dataset consists of more than 3000 light fields, which have 14$\times$14 grid angular samples. Following \cite{Tucker2020SingleViewVS,li2021mine}, we use the central 8$\times$8 gird in the experiment, where the center images of the 8$\times$8 grid are the source views and the four corner ones are the target views during testing.
}

\zy{
We carry out quantitative comparison with Tucker \etal\cite{Tucker2020SingleViewVS} and MINE\cite{li2021mine}. Since the code of \cite{Tucker2020SingleViewVS} is not publicly available, we directly compare the scores reported from their paper.
As shown in Table \ref{tab:View-Synthesis-on-Flower}, our proposed method outperforms the prior methods, indicating that stereo prior is capable to reduce the ambiguity in single-view view synthesis task and improve the performance of synthesized novel view.
\begin{table}[t]
    \centering
    \caption{\zy{Quantitative comparison of novel view synthesis on Flowers Light Fields.}}
    \begin{tabular}{c|c|c|c}
        \toprule
        \zy{Method}	& \zy{PSNR $\uparrow$} & \zy{SSIM $\uparrow$} & \zy{LPIPS $\downarrow$} \\ \midrule\midrule
        \zy{Tucker~\etal \cite{Tucker2020SingleViewVS}} &  \zy{30.1} &  \zy{0.851} &  \zy{ - }\\
        \zy{MINE \cite{li2021mine}} &  \zy{30.2} &  \zy{0.868} &  \zy{0.1603}\\
        \zy{Ours} &  \zy{\textbf{30.5}} &  \zy{\textbf{0.879}} &  \zy{\textbf{0.1582}}\\
        \bottomrule
    \end{tabular}
    \label{tab:View-Synthesis-on-Flower}\vspace{-2mm}
\end{table}
}

\zy{Besides, we showcase the qualitative comparison results in Figure \ref{fig:View-Synthesis-on-Flower}. As demonstrated in Figure \ref{fig:View-Synthesis-on-Flower}, there are some artifacts in the results produced by MINE\cite{li2021mine}. However, our proposed method achieves better photorealism. It indicates our method is capable to handle novel view beyond stereo view well.
\begin{figure}[t]
	\centering
	\setlength{\tabcolsep}{1.5pt}
	\begin{tabular}{ccc}
		\rotatebox {90}{\hspace{0.8cm}\zy{GT}} &
		\includegraphics[width=.47\linewidth, height=2.5cm]{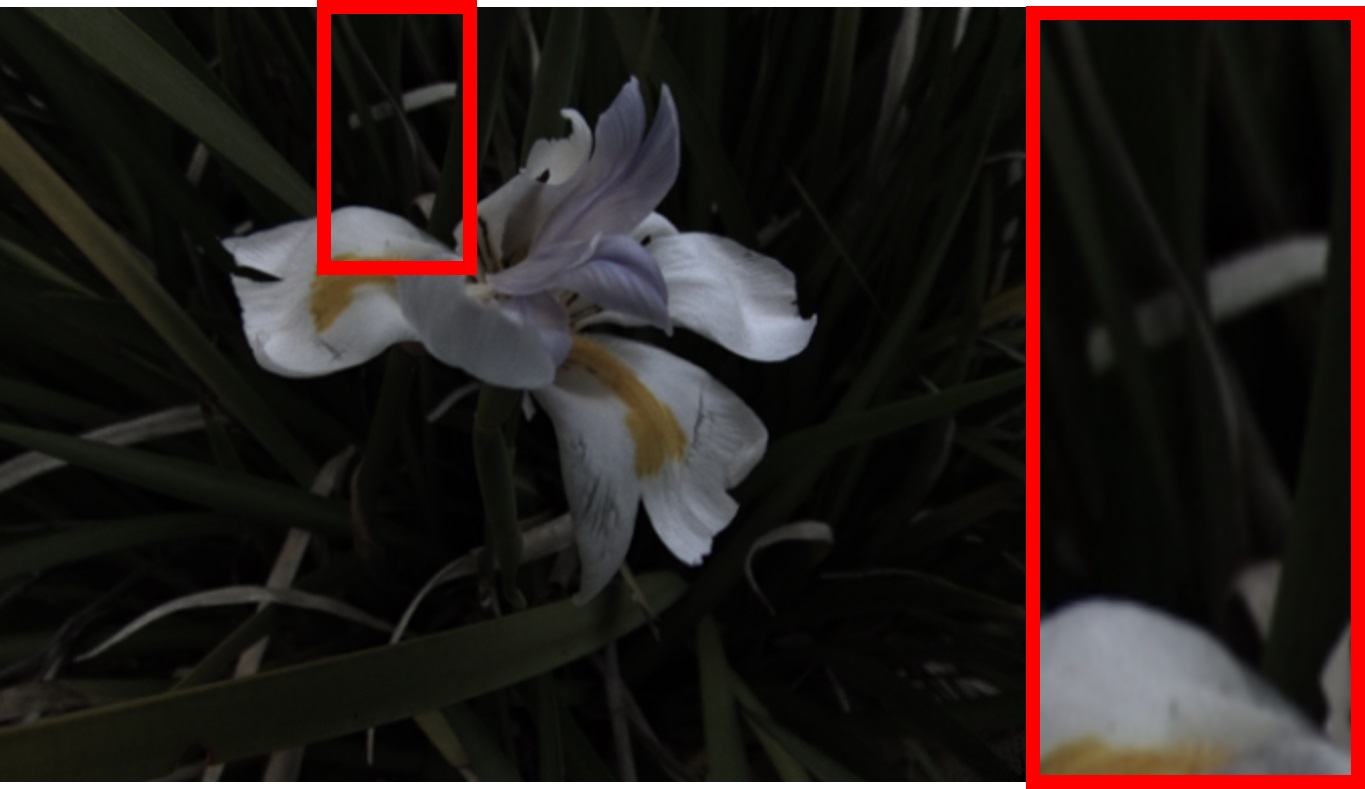} &
		\includegraphics[width=.47\linewidth, height=2.5cm]{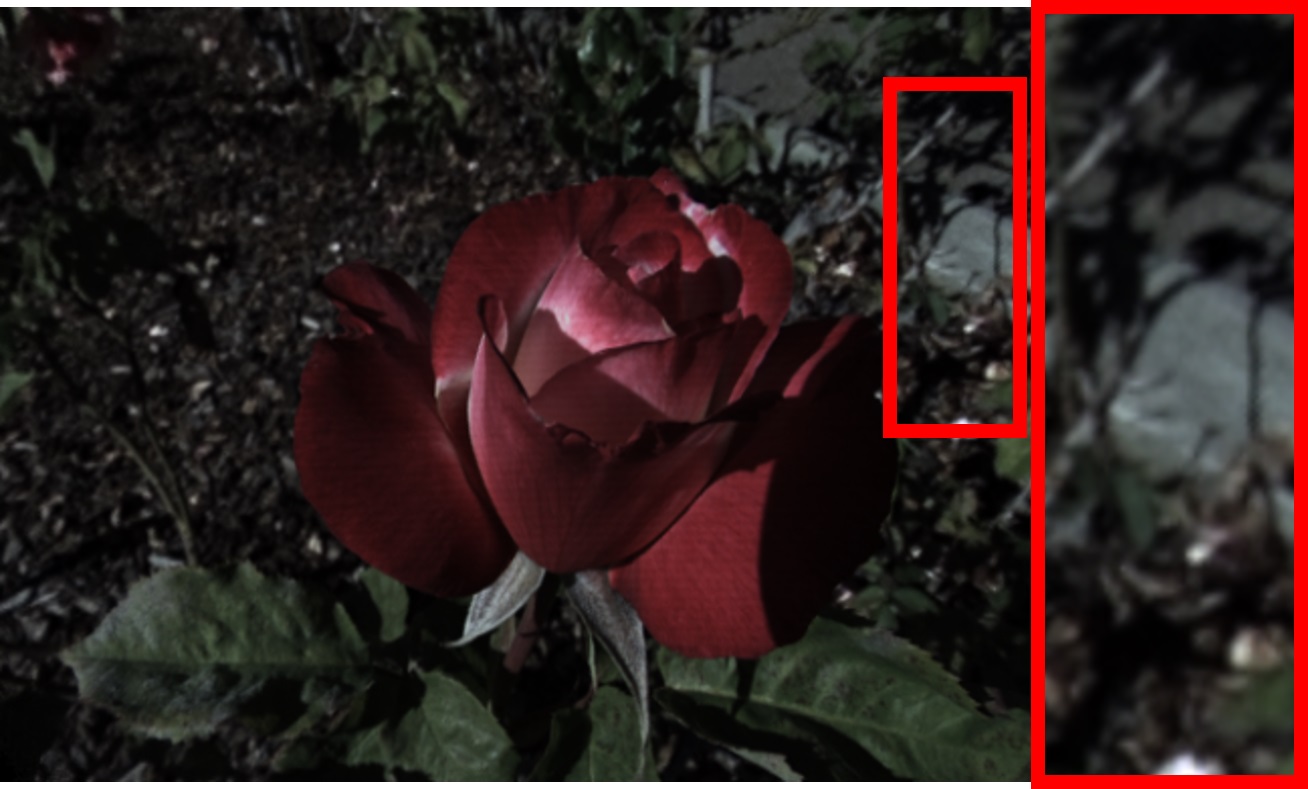} \\		
		\rotatebox {90}{\hspace{0.8cm}\zy{\cite{li2021mine}}} &
		\includegraphics[width=.47\linewidth, height=2.5cm]{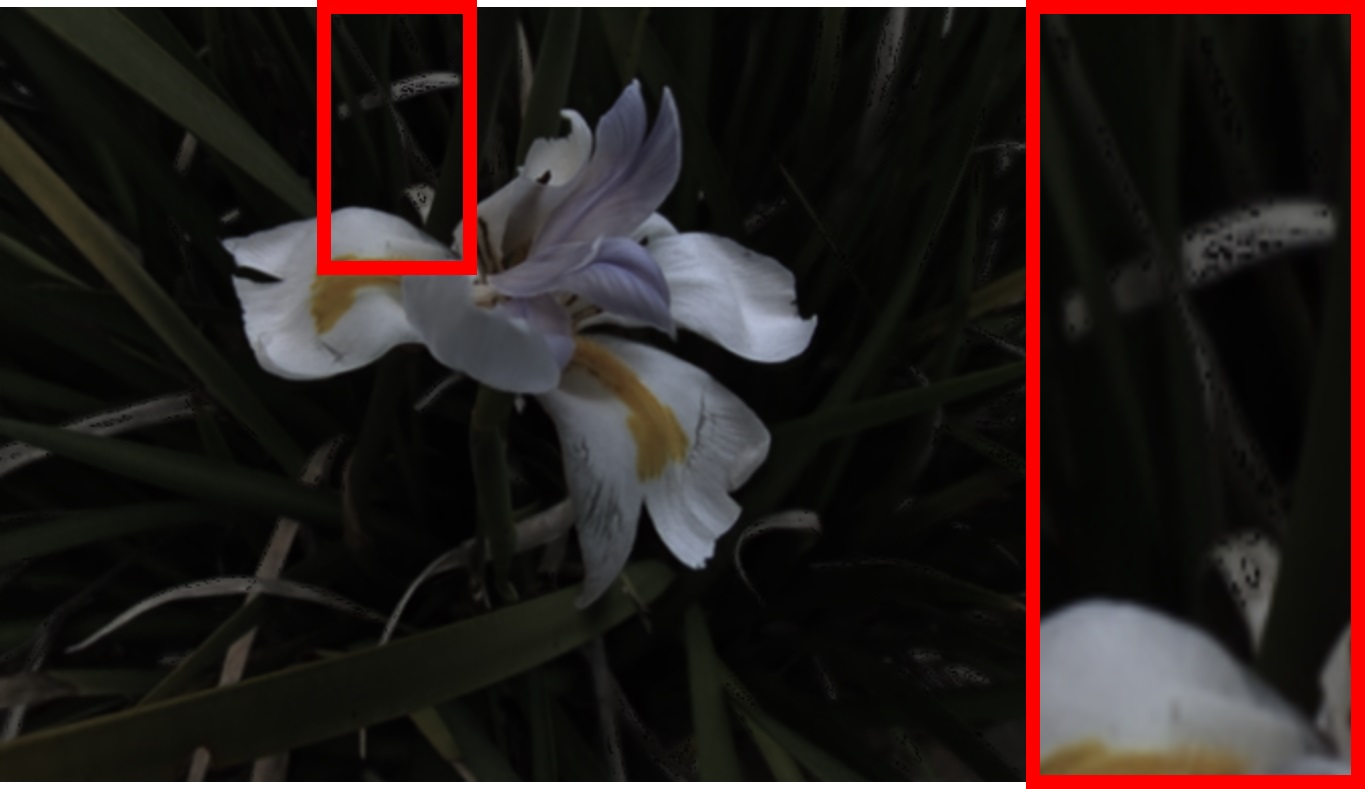} &
		\includegraphics[width=.47\linewidth, height=2.5cm]{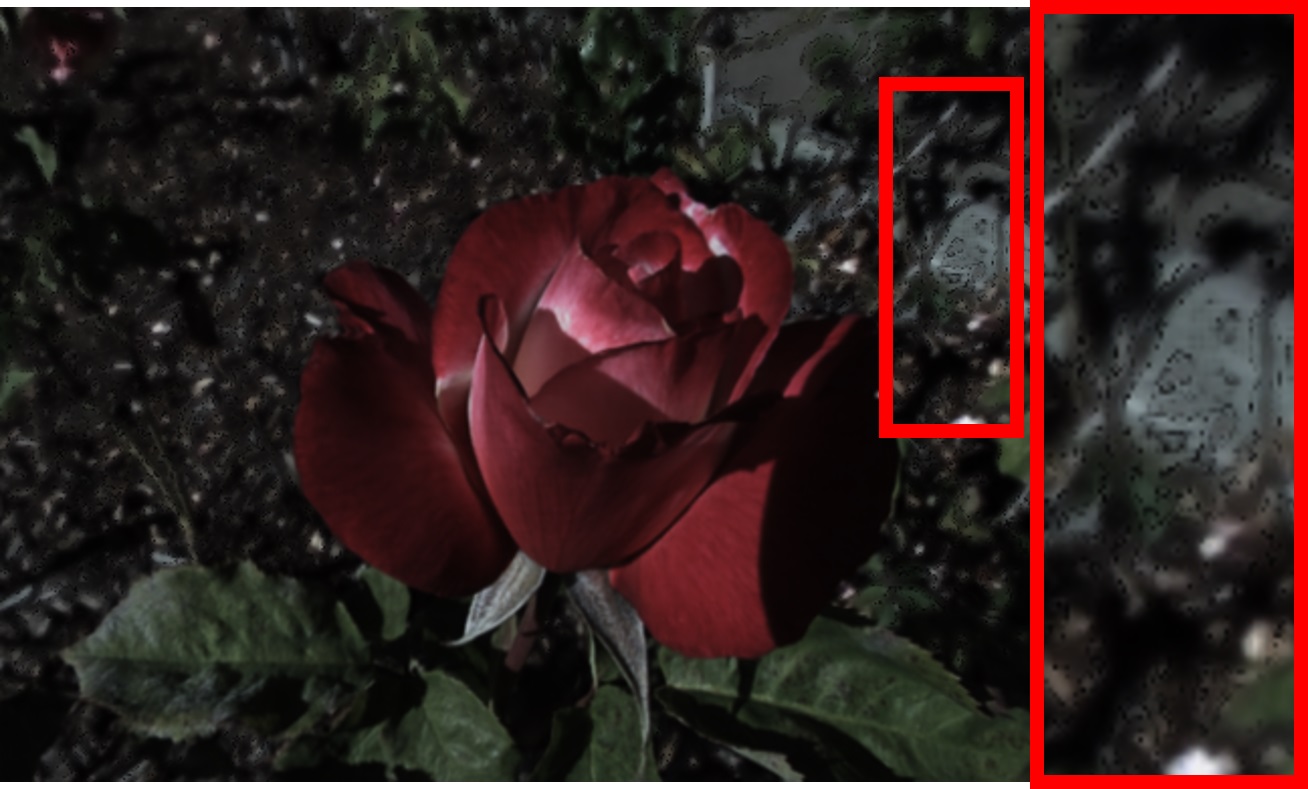} \\
		\rotatebox {90}{\hspace{0.7cm}\zy{Ours}} &
		\includegraphics[width=.47\linewidth, height=2.5cm]{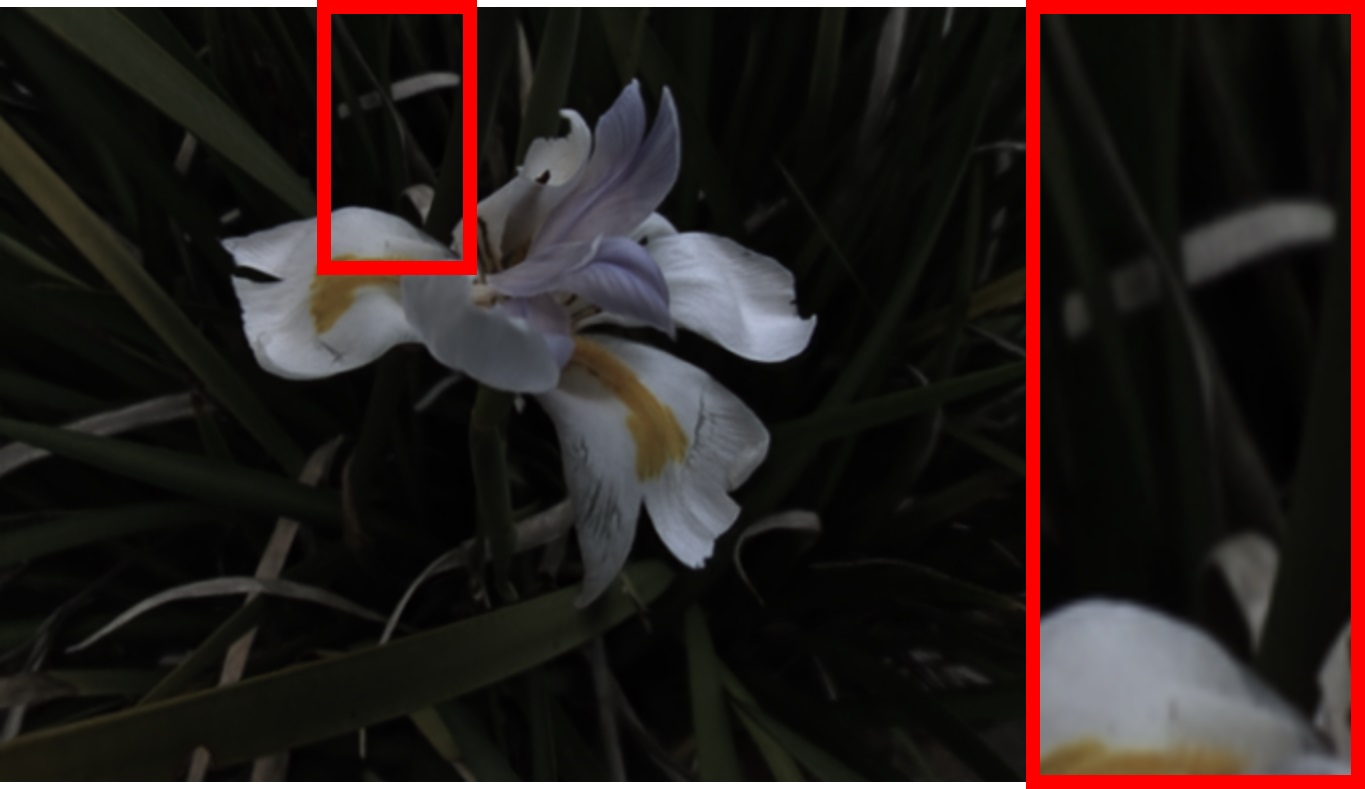} &
		\includegraphics[width=.47\linewidth, height=2.5cm]{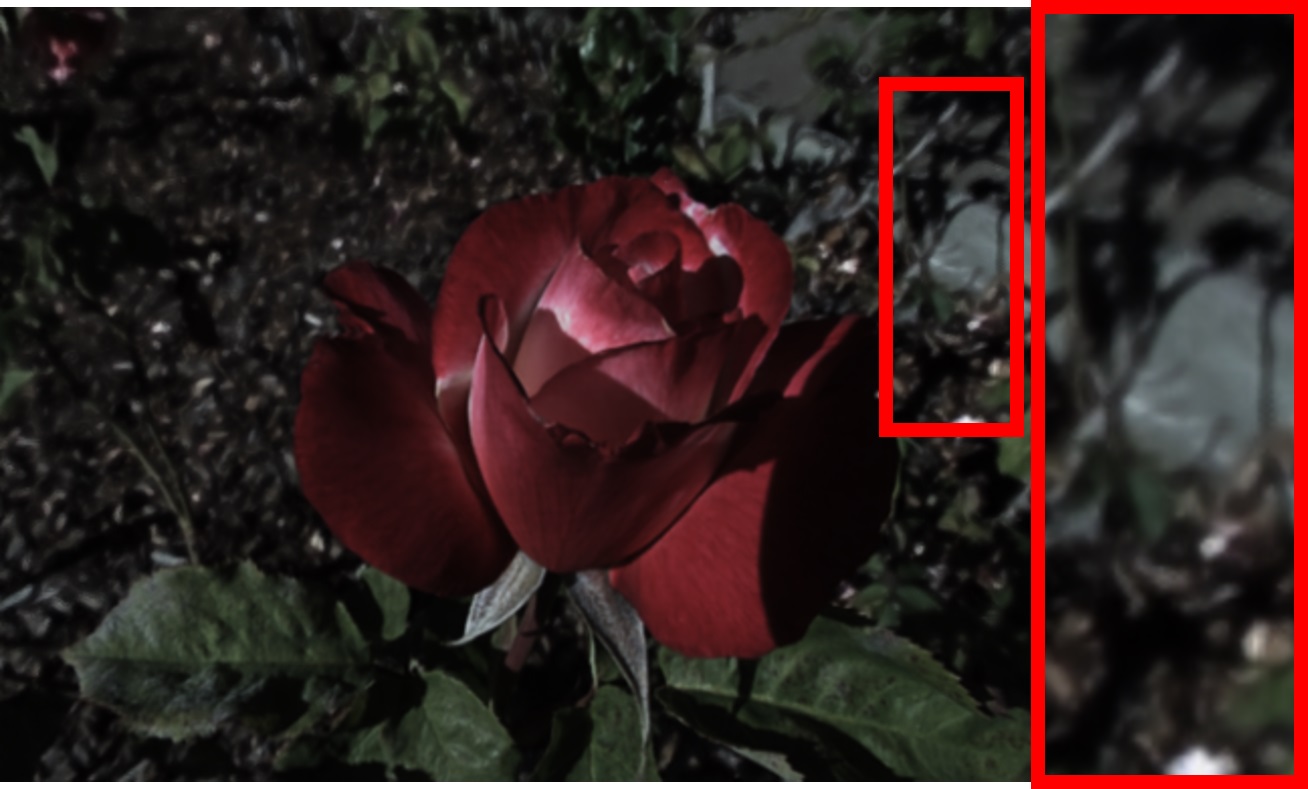}\\
	\end{tabular}
	\caption{\zy{Qualitative comparison of novel view synthesis on the Flowers light fields dataset~\cite{srinivasan2017learning}. (Better viewed in digital version with zoom function.)}}
	\vspace{-2mm}
	\label{fig:View-Synthesis-on-Flower}
\end{figure}
} 

\section{Conclusion and Limitation}
In this paper, we present a novel single-view view synthesis approach by decoupling the problem into two subproblems, pseudo stereo synthesis and 3D reconstruction. In particular, we propose self-supervised stereo warping to preserve clear details, as well as \zy{synthesis rectification} to rectify the inconsistent warped region, in order to obtain the faithful image synthesis and relieve the ambiguity for 3D reconstruction.

Both qualitative and quantitative experimental results show that our method generate more realistic contents and produce less visual artifacts than prior methods.
As for the limitation, our method now still requires stereo data for training to achieve the stereo prior, which limits our generalizability for a variety of challenging scenes. As the future work, we may explore a free viewpoint synthesis solution to resolve this concern. 

\begin{acknowledgements}
This project is supported by the National Natural Science Foundation of China (No. 61972162); Project of Strategic Importance in The Hong Kong Polytechnic University (project no. 1-ZE2Q); Guangdong Natural Science Foundation (No. 2021A1515012625); Guangdong Natural Science Funds for Distinguished Young Scholar (No. 2023B1515020097); Singapore Ministry of Education Academic Research Fund Tier 1 (MSS23C002).
\end{acknowledgements}

\bibliographystyle{spmpsci}      
\bibliography{refer}   

\end{document}